\documentclass[journal]{IEEEtran}
\usepackage{caption}
\usepackage{multirow}
\usepackage{amsmath,amssymb,amsfonts}
\usepackage{breqn}
\usepackage{algorithm}
\usepackage{graphicx}
\usepackage{textcomp}
\usepackage{multicol}
\usepackage{multirow}
\usepackage{physics}
\usepackage{upgreek}
\usepackage{xcolor}
\usepackage{caption}
\usepackage{subcaption}
\usepackage{hyperref}
\usepackage{cite}

\newcommand{\take}[1]{\textit{#1}}
\newcommand{\settings}[1]{(Settings: #1)}
\newcommand{\heading}[1]{\noindent \textbf{#1} \space}
\DeclareMathOperator*{\argmax}{argmax}
\DeclareMathOperator*{\argmin}{argmin}

\DeclareMathOperator*{\relu}{relu}

\usepackage{algpseudocode}
\algnewcommand\algorithmicinput{\textbf{Input:}}
\algnewcommand\Input{\item[\algorithmicinput]}
\algnewcommand\algorithmicoutput{\textbf{Output:}}
\algnewcommand\Output{\item[\algorithmicoutput]}

\usepackage{multirow}

\begin{document}

\title{CVPR: Consistent Valid Physically-Realizable Adversarial Attack against Crowd-flow Prediction Models}
\title{Consistent Valid Physically-Realizable Adversarial Attack against Crowd-flow Prediction Models}

\author{Hassan Ali$^1$, Muhammad Atif Butt$^1$, Fethi Filali$^{2}$, Ala Al-Fuqaha$^{3}$, and Junaid Qadir$^{4}$\thanks{Corresponding author: Junaid Qadir (junaid.qadir@qu.edu.qa) }\\
$^1$ Information Technology University (ITU), Punjab, Lahore, Pakistan\\
$^2$ Qatar Mobility Innovations Center (QMIC), Doha, Qatar \\
$^3$ Hamad Bin Khalifa University (HBKU), Doha, Qatar \\
$^4$ Qatar University, Doha, Qatar
}

\maketitle

\begin{abstract}
Recent works have shown that deep learning~(DL) models can effectively learn city-wide crowd-flow patterns, which can be used for more effective urban planning and smart city management. However, DL models have been known to perform poorly on inconspicuous adversarial perturbations. Although many works have studied these adversarial perturbations in general, the adversarial vulnerabilities of deep crowd-flow prediction models in particular have remained largely unexplored. In this paper, we perform a rigorous analysis of the adversarial vulnerabilities of DL-based crowd-flow prediction models under multiple threat settings, making three-fold contributions. (1) We propose \emph{CaV-detect} by formally identifying two novel properties---\underline{C}onsistency \underline{a}nd \underline{V}alidity---of the crowd-flow prediction inputs that enable the \underline{detect}ion of standard adversarial inputs with 0\% false acceptance rate (FAR). (2) We leverage universal adversarial perturbations and an adaptive adversarial loss to present adaptive adversarial attacks to evade \emph{CaV-detect} defense. (3) We propose \emph{CVPR}, a \underline{C}onsistent, \underline{V}alid and \underline{P}hysically-\underline{R}ealizable adversarial attack, that explicitly inducts the consistency and validity priors in the perturbation generation mechanism. We find out that although the crowd-flow models are vulnerable to adversarial perturbations, it is extremely challenging to simulate these perturbations in physical settings, notably when \emph{CaV-detect} is in place. We also show that \emph{CVPR} attack considerably outperforms the adaptively modified standard attacks in FAR and adversarial loss metrics. We conclude with useful insights emerging from our work and highlight promising future research directions.
\end{abstract}


\IEEEpeerreviewmaketitle

\section{Introduction}
The crowd-flow prediction problem aims to predict the crowd-flow state at some future time, given a set of crowd-flow states at previous times. Crowd-flow prediction models can be profitably used in diverse fields including modeling and understanding human behavior~\cite{saadatnejad2022socially}; transportation management~\cite{zhang2017deep}; and smart-city planning~\cite{jiang2022taxibj21}. Particularly, for smart city settings, crowd-flow prediction models can be critical in identifying over-crowded regions and be used for adopting appropriate preemptive actions to ensure human safety. 

Deep Neural Networks~(DNNs) represent a promising technique for solving the crowd-prediction problem, having shown their versatility on many complex practical tasks~\cite{petrick2021spie, ali2021all, ali2022tamp, khalid2020fadec, butt2023towards}.
Recent works have successfully leveraged DNNs to not only predict the future crowd-flow states~\cite{zhang2017deep,jiang2022taxibj21, li2017diffusion} but also for related tasks such as predicting pedestrian trajectory~\cite{saadatnejad2022socially} and a vehicle's travel time. However, the performance of a DNN highly depends on its training data, which causes it to be vulnerable to adversarial perturbations---undetectable noise, intentionally induced in the input in order to change the DNN output~\cite{khalid2020fadec,ilyas2019adversarial,szegedy2014intriguing}. Although several crowd-flow prediction models based on diverse architectures have been proposed, to the best of our knowledge, the adversarial vulnerabilities of these models remain largely unexplored.

\begin{figure}
    \centering
    \includegraphics[width=0.5\linewidth, page=1]{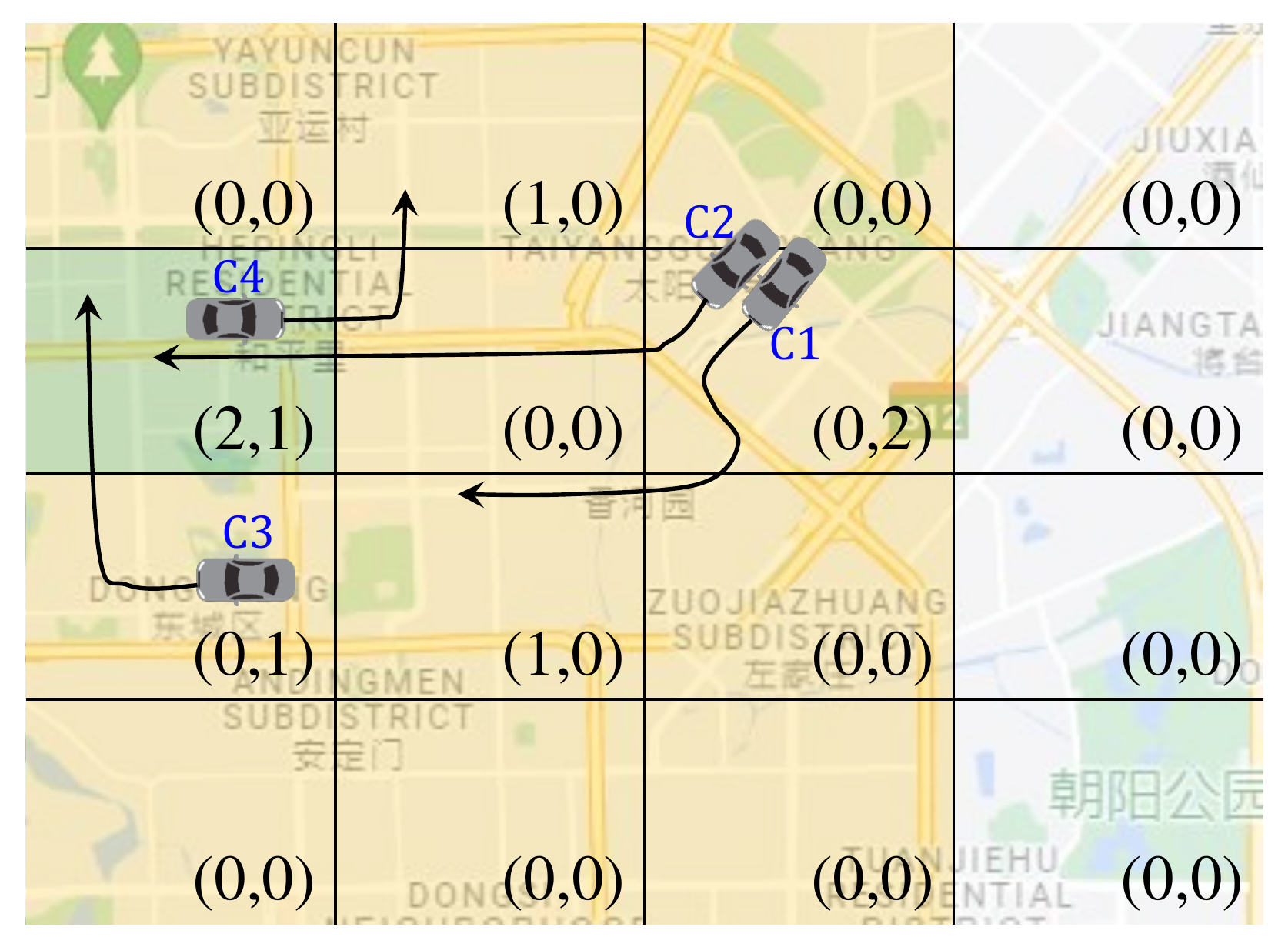}
    \caption{An illustration of gridding a region and computing the inflow and the outflow matrices from the flow of crowd between adjacent regions (grid points). We typically assume the adjacency within the $2^{nd}$ neighborhood---the adjacent grid points of green highlighted area are highlighted yellow.}
    \label{fig:inflow_outflow_illustration}
\end{figure}

In this paper, we bridge this gap by studying the worst-case performance of three popular and diverse crowd-flow prediction models---Multi-Layer Perceptron (MLP)~\cite{jiang2022taxibj21}, Spatio-Temporal Resnet (STResnet)~\cite{zhang2017deep} and Temporal Graph Convolutional Neural Network (TGCN)~\cite{zhao2019t}---under multiple attack settings, including realistic physical scenarios. For evaluation, we consider the TaxiBJ dataset, which is one of the most commonly used datasets for crowd-flow prediction. TaxiBJ divides a city into $32 \times 32$ grid points (regions) and records the region-wise crowd inflow\footnote{Total devices flowing into a grid point from its adjacent grid points.} and outflow\footnote{Total devices flowing out of a grid point into its adjacent grid points.} at half-hourly intervals~\cite{mourad2019astir, jiang2022taxibj21, zhang2017deep}, as illustrated in Fig.~\ref{fig:inflow_outflow_illustration}.

\begin{figure}
    \centering
    \begin{subfigure}[t]{0.485\linewidth}
        \centering
        \includegraphics[width=1\linewidth, page=2]{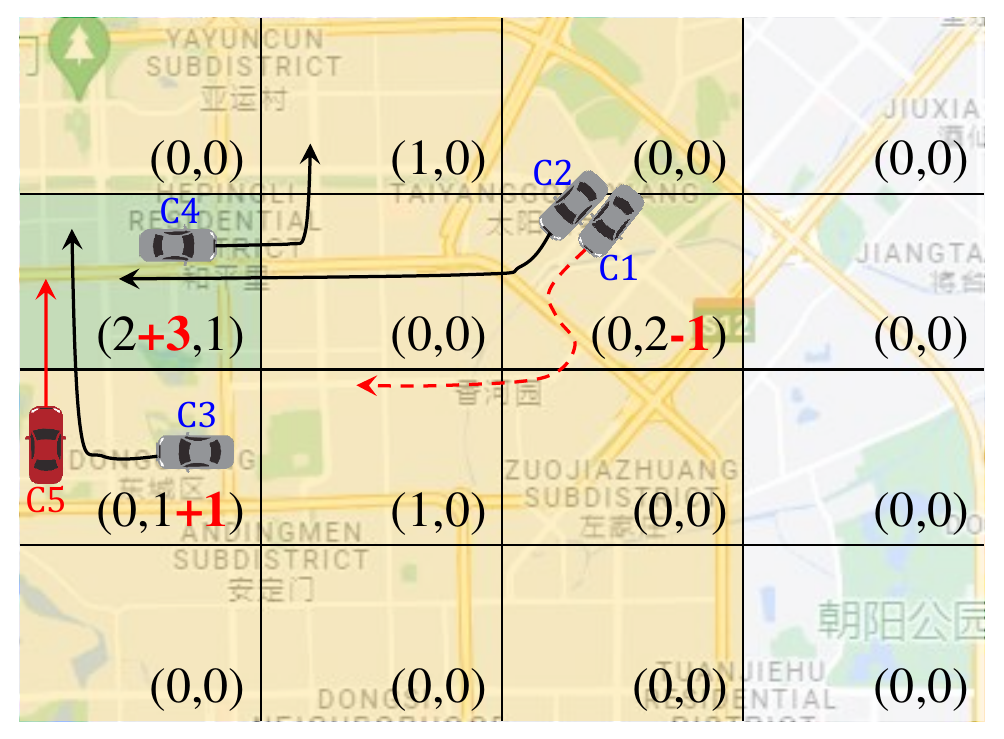}
        \caption{Perturbed by PGD attack}
    \end{subfigure}
    \begin{subfigure}[t]{0.485\linewidth}
        \centering
        \includegraphics[width=1\linewidth, page=3]{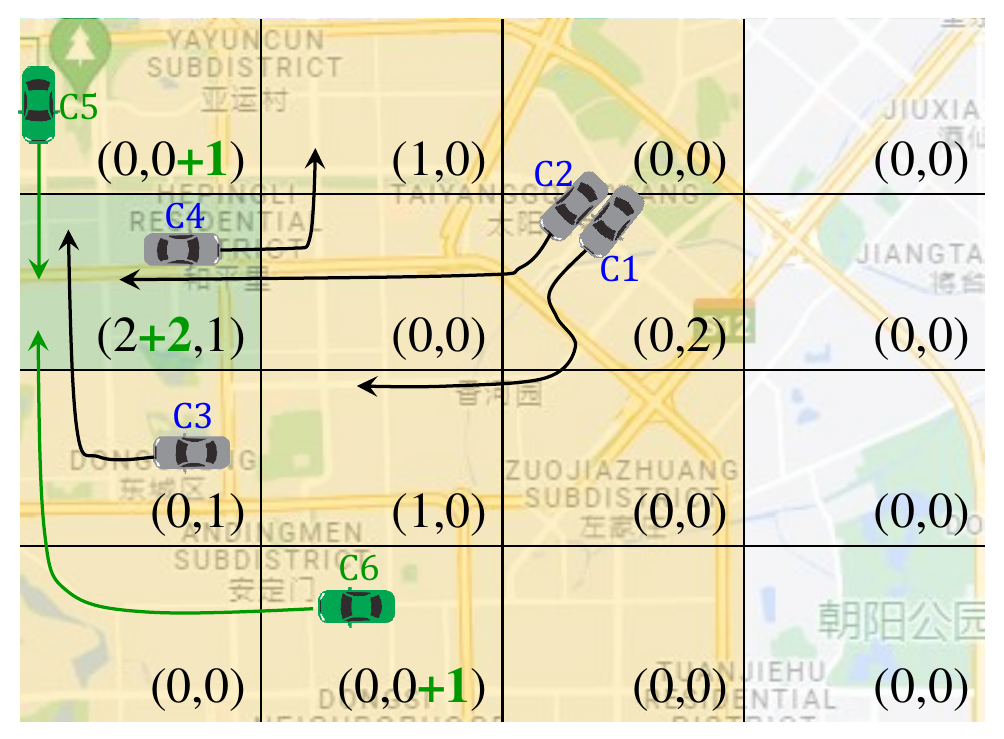}
        \caption{Perturbed by \emph{CVPR} attack}
    \end{subfigure}
    
    \caption{An illustration of \emph{invalid} and \emph{valid} adversarial inputs generated by the standard PGD attack and our proposed \emph{CVPR} attack respectively. In (a), for the grid point-$(1,0)$ highlighted green, the total (perturbed) inflow recorded is five. Three of the five inflowing devices can be C2, C3, C5 outflowing from the adjacent regions (highlighted yellow). Where are the other two devices outflowing from? In (b), \emph{CVPR} attack perturbs the outflow of adjacent regions based on the inflow perturbations of the grid point-$(1,0)$ for physical plausibility.}
    \label{fig:validity_illustration}
\end{figure}

\vspace{1mm}
\heading{Challenges:} Firstly, the structure of inputs to different crowd-flow prediction models vary significantly, and therefore, the adversarial evaluation results may not fairly compare different models. For example, the TGCN~\cite{zhao2019t} only expects the crowd-flow state history of a pre-defined length at half-hourly intervals as input. In contrast, STResnet~\cite{zhang2017deep} takes three sets of inputs representing the hourly, the daily, and the weekly history of the pre-defined lengths. In order to fairly evaluate the robustness of features enabled by different architectures, the models should be evaluated under similar input settings.

Secondly, recent years have witnessed an arms race between the attackers trying to fool a DNN and the defenders trying to defend against these attacks---most of the attacks and defenses were proven ineffective by more adaptive defenses and attacks, respectively, within a few months after they were proposed. Therefore, developing novel adaptive defense and attack strategies to comprehensively analyze adversarial attacks (and their limitations) on DNNs is both extremely important and challenging~\cite{athalye2018obfuscated, tramer2020adaptive}.

\vspace{1mm}
\heading{Findings and Contributions:} We first analyze different crowd-flow prediction models against three standard adversarial attacks---FGSM, i-FGSM, and PGD attacks---and show that the crowd-flow prediction models, much like other deep learning models, are significantly vulnerable to the adversarial attacks, under several design choices. However, we note that these vulnerabilities are mainly limited to the digital attack setting, which assumes a worst-case attacker who has access to the digital input pipeline of the crowd-flow prediction model.

We then identify two properties---\emph{consistency} and \emph{validity}---that natural crowd-flow inputs must satisfy. Although these properties are natural and intuitive, to the best of our knowledge, they have not been emphasized or used in previous works. The property of \textit{consistency} requires that the crowd-flow state history at some time, $t$, must be consistent with the crowd-flow states at the previous times.  In relation to \textit{validity}, the inflow to and outflow from a particular grid point at any given time, by definition, must always be less than the accumulative outflows from and inflows to the adjacent grid points respectively. As illustrated in Fig.~\ref{fig:validity_illustration}(a) with example, adversarial perturbations of standard attacks contradict these relationships, and therefore, can be easily invalidated at test time. Noting that the adversarial inputs generated by the standard attacks are \textit{inconsistent} and \textit{invalid}, we show the usefulness of these properties by proposing \textbf{\emph{CaV-detect}}, a novel consistency and validity check mechanism to detect adversarial inputs at test time by analyzing the inflow and outflow matrices (to check the validity) and comparing the crowd-flow state history (to ensure consistency). Results show that \emph{CaV-detect} can detect standard adversarial inputs with 0\% FAR (FRR $\leq$ 0.5\%).

Assuming an expert attacker, we adaptively modify standard adversarial attacks to evade \emph{CaV-detect} by combining universal adversarial perturbations~\cite{moosavi2017universal} and adaptive adversarial loss. Compared to non-adaptive standard attacks with FAR of 0\%, the adaptive attacks typically achieved FAR of $\geq$80\% (FRR $\leq$ 0.5\%). 

We then propose \textbf{\emph{CVPR} attack, a \underline{C}onsistent, \underline{V}alid, and \underline{P}hysically-\underline{R}ealizable} adversarial attack that explicitly inducts the consistency and validity priors in the adversarial input generation mechanism to find consistent and valid adversarial perturbations (see Fig.~\ref{fig:validity_illustration}(b)), and outperforms the standard and the adaptive attacks in both the FAR ($\approx$100\%) and the adversarial loss against \emph{CaV-detect}.

We also note that the perturbations generated by the standard adversarial attacks are often negative, and thus, cannot be realized by a physical adversary---an adversary who can only control a limited number of physical devices in the city, and has no access to the digital input pipeline. Therefore, we modify the adversarial loss to propose and evaluate a physically realizable attack by generating adversarial perturbations that can be physically simulated in the real world. Our findings highlight that realizing the adversarial perturbations under the physical setting requires an impractically large number of adversarially controlled devices, particularly, when \emph{CaV-detect} is in place.

Finally, our qualitative evaluations show that the crowd-flow prediction models exhibit limited expressiveness---the resulting models, despite showing small test errors, are incapable of producing certain outputs. We attribute this to TaxiBJ data comprising clustered and highly similar crowd-flow states~\cite{jiang2022taxibj21}.

Our main contributions are listed below:
\begin{itemize}
    \item We are the first to study the adversarial vulnerabilities of the crowd-flow prediction models.
    \item We formalize two novel properties ---\textit{consistency} and \textit{validity}---of crowd-flow prediction inputs and show their usefulness by proposing a novel defense method named \emph{CaV-detect} that achieves 0\% FAR with $\leq$0.5\% false rejection rate (FRR).
    \item We combine adaptive loss with universal adversarial perturbation to exhaustively test \emph{CaV-detect}.
    \item We induct the consistency and validity priors in the adversarial input generation mechanism to propose \emph{CVPR} attack that addresses several shortcomings of the standard attacks.
\end{itemize}

\section{Related Work}
Owing to the recent developments in intelligent transportation systems (ITS), road traffic congestion forecasting is becoming one of the key steps in curtailing delays and associated costs in traffic management \cite{chen2018pcnn}. In the following, we highlight some of the notable and recent works on crowd-flow state prediction.

\subsection{Crowd-flow State Prediction}
Depending on the characteristics, structure and quality of the data, various kinds of machine learning  (ML) techniques are employed to develop road traffic congestion models. In the literature, these crowd-flow prediction techniques are widely categorized into three main branches---probabilistic and statistical reasoning-based crowd-flow models~\cite{onieva2012genetic, zhang2014hierarchical, onieva2016comparative, lopez2015hybrid, zheng2018real, zaki2020traffic, sun2019traffic, qi2014hidden, yang2013feature, zhu2019early, sun2006bayesian, asencio2016novel, kim2016diagnosis}, shallow ML techniques~\cite{yang2019application, nadeem2018performance, lee2015prediction, zhang2018user, jain2017traffic, tseng2018congestion, wang2015short, liu2017prediction, chen2019discrimination} and deep learning (DL) models\cite{ma2017learning, chen2018pcnn, ma2015large, sun2019traffic, xing2019large, lin2017interval}. Our work focuses on studying the adversarial vulnerabilities of DL-based crowd-flow state prediction models. More specifically, we choose three crowd-flow prediction models of notably different architectures for the robustness evaluation. Our choices are based on the recency, diversity, relevancy to the problem, and popularity of the architecture. All of these model architectures were trained and evaluated on TaxiBJ dataset in their original papers.

\vspace{1mm}
\heading{MLP architecture.} In their recent work, Jiang et al.~\cite{jiang2022taxibj21} present TaxiBJ dataset for the year 2021, and use a simple MLP model to benchmark their results. We choose the MLP model motivated by its recency, simplicity, and adversarial transferability---recent works have shown that compared to other architectures, adversarial inputs generated against the MLP models are comparatively general and more effectively transfer to different architectures~\cite{ali2021all}.

\vspace{1mm}
\heading{STResnet architecture.} STResnet model proposed by Zhang et al.~\cite{zhang2017deep} is built over the spatio-temporal residual unit modeling both the spatial dependencies using convolutional layers and the temporal dependencies by concatenating crowd-flow states from the recent past into a tuple. We choose this model motivated by its highly relevant architecture (of spatio-temporal nature) and popularity (1425 citations\footnote{scholar.google.com, 3 November 2022}). However, the model proposed in \cite{zhang2017deep} is trained over hourly, daily, and weekly history concatenated to form a single input tuple. For a fair comparison of different architectures, we modify the architecture to match our input structure, common for all architectures, by training it only over the half-hourly history.

\vspace{1mm}
\heading{T-GCN architecture.} When modelling crowd-flow patterns, graph convolutional networks (GCN) are a gold-standard choice. Zhao et al.~\cite{zhao2019t} propose a Temporal Graph Convolutional Network (TGCN) that exploits temporally updated convolutional graph networks to model both the temporal and the spatial dependencies in the input. We choose the proposed TGCN model motivated by its relevance (spatio-temporal architecture) and popularity (839 citations\footnote{scholar.google.com, 3 November 2022}).


\subsection{Adversarial Attacks on DL Models}
Adversarial attacks are small imperceptible changes to the input to fool DNNs. Since they were discovered by Szegedy et al.~\cite{szegedy2013intriguing} for image classification, many works have shown that DNNs are generally vulnerable to these attacks in a range of applications including Computer Vision (CV)~\cite{khalid2020fadec, butt2023towards}, audio processing~\cite{latif2018adversarial}, networking~\cite{usama2019black, usama2018adversarial, usama2019generative} and Natual Language Processing(NLP)~\cite{ali2021all, morris2020textattack}. Adversarial attacks may assume a black-box~\cite{chen2017zoo, khalid2020fadec, brendel2017decision} or a white-box threat model~\cite{goodfellow2014explaining, madry2017towards, carlini2017towards}. Despite several efforts to defend against these attacks~\cite{goodfellow2014explaining, khalid2019qusecnets, ali2019sscnets, papernot2016distillation, dhillon2018stochastic}, certified defenses~\cite{huang2021training} and adversarial training~\cite{madry2017towards} are the only effective defenses that provide reliable robustness to DL models~\cite{lee2021towards, athalye2018obfuscated}.

In this work, we assume a white-box threat model assuming an attacker knowledgeable of the model architecture and weights. Let an input $x \in \mathcal{X}$, where $\mathcal{X}$ denotes the valid input feature space, produce a true output $\mathcal{F}_\theta(x)$, where $\theta$ denotes the learnable parameters of $\mathcal{F}$. The goal of an attack is to compute an adversarial perturbation $\delta^*$, in order to get the desired output, $y_{target}$ from the model when the perturbation is added to the input.

\begin{equation}
    \delta^* = \argmin_{\delta \in \mathcal{B}(\epsilon)} (\mathcal{F}_{\theta}(x + \delta)-y_{target})^2
    \label{eq:general_adversarial_goal}
\end{equation}
where $\mathcal{B}(\epsilon)$ denotes a pre-defined bounded set of allowed perturbations. One of the most common choices for $\mathcal{B}(\epsilon)$ is an $l_\infty$ ball, defined as, ${\boldsymbol{\updelta} \in \mathcal{B}_\infty(\epsilon) := -\epsilon \leq \boldsymbol{\updelta} \leq \epsilon}$. Eq-\eqref{eq:blind_attack} is iteratively optimized depending on the attack algorithm~\cite{szegedy2014intriguing}.

\vspace{1mm}
\textbf{Fast Gradient Sign Method}~(FGSM)~\cite{goodfellow2014explaining}: FGSM attack achieves eq-\eqref{eq:general_adversarial_goal} by computing the adversarial perturbation, $\delta^*$, in a single step as,
\begin{equation}
    \delta^* = \delta - \epsilon \times \operatorname{sign} \left( \frac{\partial (\mathcal{F}_{\theta}(x + \delta)-y_{target})^2}{\partial \delta} \right)
\end{equation}

\vspace{1mm}
\textbf{Iterative-FGSM}: i-FGSM attack achieves eq-\eqref{eq:general_adversarial_goal} by computing the adversarial perturbation, $\delta^*$, in $N$ steps as,
\begin{align}
    \delta_{i+1} = \delta_i - \frac{\epsilon}{N} \times \operatorname{sign} \left( \frac{\partial (\mathcal{F}_{\theta}(x + \delta)-y_{target})^2}{\partial \delta_i} \right) \nonumber \\
    \delta_{i+1} = \operatorname{clip} (\delta_{i+1}, \epsilon, \epsilon )
    \label{eq:ifgsm_optimization}
\end{align}
where $N$ is the number of iterations, $\delta_0 = \delta$, and $\delta^* = \delta_N$.

\vspace{1mm}
\textbf{Projected Gradient Descent}~(PGD)~\cite{madry2017towards}: The optimization in eq-\eqref{eq:general_adversarial_goal} is achieved by repeating the steps in eq-\eqref{eq:pgd_optimization} for $[0,N-1]$ steps,

\begin{align}
    \delta_{i+1} = \delta_i - \alpha \times \operatorname{sign} \left( \frac{\partial (\mathcal{F}_{\theta}(x + \delta)-y_{target})^2}{\partial \delta_i} \right) \nonumber \\
    \delta_{i+1} = \operatorname{clip} (\delta_{i+1}, \epsilon, \epsilon )
    \label{eq:pgd_optimization}
\end{align}

where $\delta_0 = \delta$, $\delta^* = \delta_N$, and $\alpha$ denotes the PGD-step. Note that $\alpha$ is the tunable hyper-parameter of the PGD attack.

\section{Methodology}
We first formulate the crowd-flow prediction problem in the context of the TaxiBJ dataset and formally define the consistency and validity properties of crowd-flow state inputs. Based on these properties, we propose \emph{CaV-detect}, to detect adversarially perturbed inputs by analyzing their consistency and validity. Finally, we present our novel algorithm of \emph{CVPR} attack for generating consistent, valid, and physically realizable adversarial perturbations.

\begin{figure}
    \centering
    \includegraphics[width=1\linewidth]{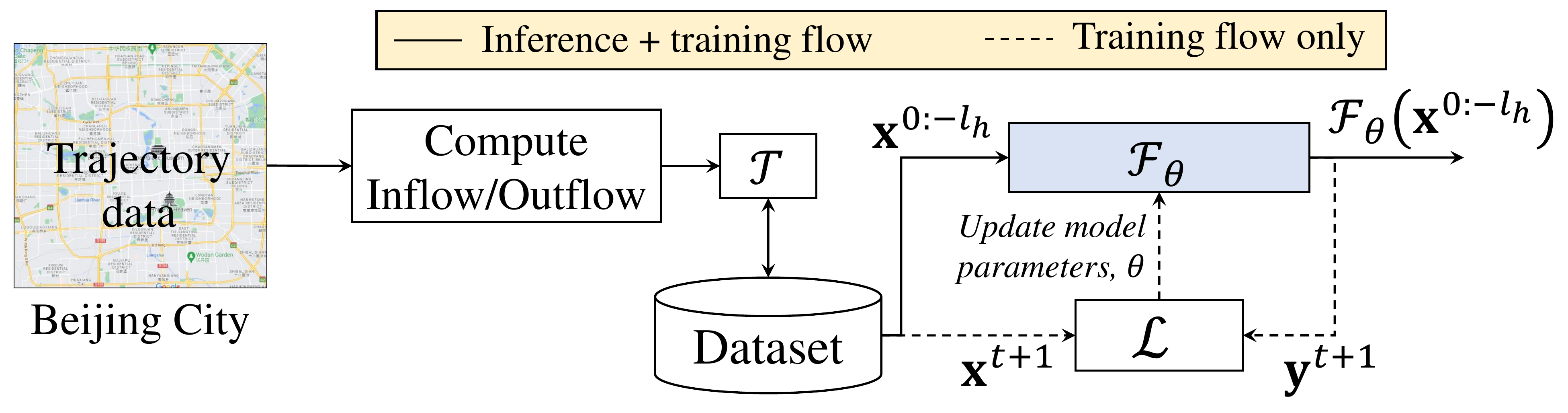}
    \caption{Illustrating the training setup of the crowd-flow prediction models for the TaxiBJ dataset. The trajectory data collected from the city is first converted into the inflow/outflow matrices and transformed using $\mathcal{T}$, which are then saved in the memory and concatenated with the history set to form a tuple input, $\vb{X}_h(t) = \bigcup_{i=0}^{h} \vb{x}^{t-i}$, to the crowd-flow prediction model.}
    \label{fig:crowdflow_prediction}
\end{figure}

\subsection{Crowd-flow State Prediction}

\vspace{1mm}
\heading{Problem formulation.}
The openly available TaxiBJ crowd-flow dataset~\cite{jiang2022taxibj21} that we use in this work divides the city into a 2-D grid of size $l_1 \times l_2$, where each grid point physically spans an area of 1000 meters square. The dataset reports the city-wide flow of the crowd as a tuple of inflow and outflow matrices after each 30 minutes interval. At any given time, $t$, the integer crowd-flow state, $\vb{n}^t \in \mathbb{Z}^{2 \times l_1 \times l_2}$, is defined as a tuple of the inflow and outflow matrices, denoted $\vb{n}^t_{in} \in \mathbb{Z}^{l_1 \times l_2}$ and $\vb{n}^t_{out} \in \mathbb{Z}^{l_1 \times l_2}$, defining the number of devices ($\approx$~persons~\cite{zhang2017deep}) flowing into and out of the grid points in the $l_1 \times l_2$ city grid, respectively. Formally,
\begin{equation}
    \vb{n}^t = (\vb{n}^t_{in}, \vb{n}^t_{out}) 
\end{equation}

We define the crowd-flow state, $\vb{x}^t \in \mathbb{R}^{2 \times l_1 \times l_2}$, as,
\begin{equation}
    \vb{x}^t = (\vb{x}^t_{in}, \vb{x}^t_{out}) = \mathcal{T}(\vb{n}^t) = (\mathcal{T}(\vb{n}^t_{in}), \mathcal{T}(\vb{n}^t_{out}))
    \label{eq:taxibj_data}
\end{equation}

where $\mathcal{T}(\cdot)$ denotes an element-wise (somewhat) reversible transformation function. A standard practice is to choose $\mathcal{T}$, such that $\forall \vb{n}^t \in [0..\infty], \vb{x}^t \in [0,1]$. Following the prior arts~\cite{zhang2017deep,jiang2022taxibj21}, we use $\mathcal{T}(\vb{n}^t) = \min(\vb{n}^t/1000, 1)$ in our experiments.

Our goal is to learn a model, $\mathcal{F}_\theta$, that predicts the crowd-flow state in the immediate future, $t+1$, given the current and the previous states, $\bigcup_{i=0}^{h} \vb{x}^{t-i}$, denoted as $\vb{X}_h(t)$ in future, where $h$ is the history length denoting the total number of previous crowd-flow states concatenated together with the current state as a tuple input to $\mathcal{F}_\theta$. Formally,

\begin{equation}
    \vb{y}^{t+1} = \mathcal{F}_\theta(\vb{X}_h(t))
    \label{eq:regression_problem}
\end{equation}

where $\vb{y}^{t+1}$ denotes the output of the model. Similar to the previous studies, we solve the above problem as a regression task to learn a model $\mathcal{F}_{\theta^*}$ that minimizes the expectation of the squared difference of the outputs and the ground truths over the whole TaxiBJ dataset, $\mathcal{D}$,

\begin{equation}
    \theta^* = \argmin_{\theta} \mathbb{E}_{x \sim \mathcal{D}}[(\mathcal{F}_\theta(\vb{X}_h(t)) - \vb{x}^{t+1})^2]
    \label{eq:learning_task}
\end{equation}

The training setup that we use for training the crowd-flow prediction models, $\mathcal{F}_{\theta}$, is shown in Fig.~\ref{fig:crowdflow_prediction}.




\begin{figure*}
    \centering
    \includegraphics[width=1\linewidth]{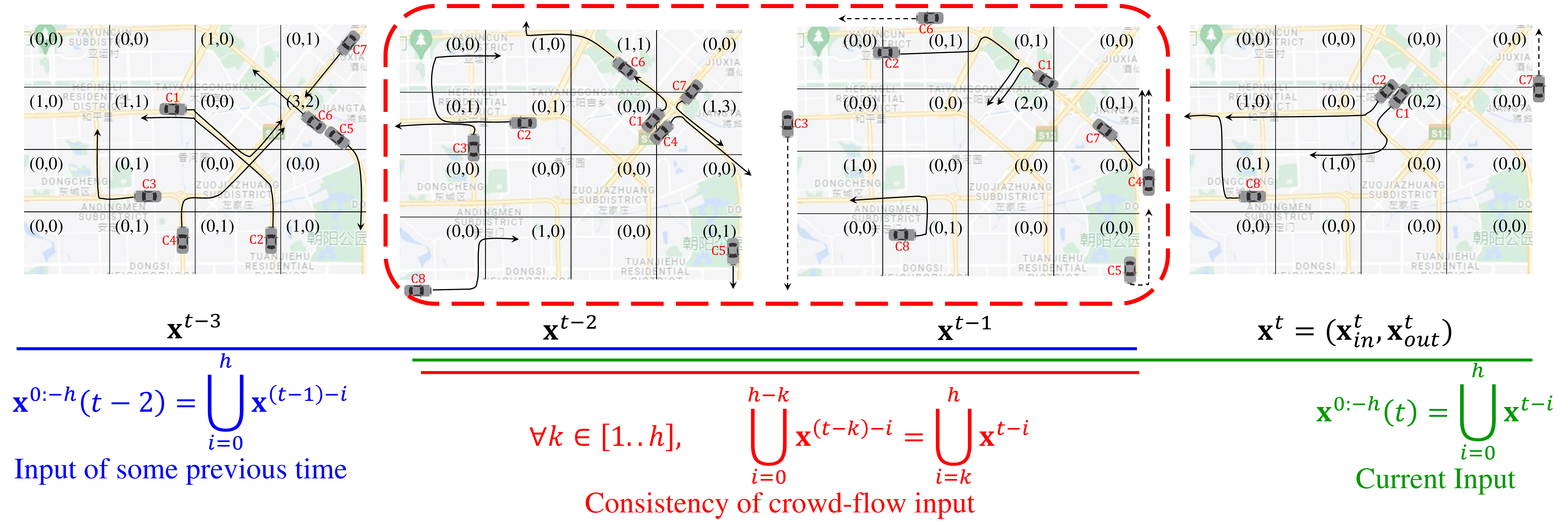}
    \caption{An illustration of the \emph{consistency} property of crowd-flow state inputs. The crowd-flow state history at any time $t$, must be consistent with the crowd-flow states recorded at the previous times.}
    \label{fig:consistency_illustration}
\end{figure*}

\vspace{1mm}
\heading{Properties of the crowd-flow state inputs.}
Here we formally define two key properties of crowd-flow prediction inputs, consistency and validity, which enable the development of \emph{CaV-detect}. We also formally analyze eq-\eqref{eq:general_adversarial_goal} (in specific regards to the aforementioned properties) to highlight the limitations of adversarial attacks against the crowd-flow prediction inputs.

\begin{enumerate}
    \item \heading{Consistency:} 
    We consider a sequence of crowd-flow states, $\bigcup_{i=0}^{-2h} \vb{x}^{t-i}$, recorded at different time intervals from ${t-2h}$ to $t$,
    \begin{equation}
        \bigcup_{i=0}^{2h} \vb{x}^{t-i} = \{\vb{x}^{t-2h}, \dots, \vb{x}^{t-h}, \dots, \vb{x}^t\}
    \end{equation}
    where $h$ denotes the history length and $\vb{x}^t$ denotes the crowd-flow state at time, $t$. The data preprocessing step prepares a history set containing $h$ previous crowd-flow states, $\bigcup_{i=1}^{h} \vb{x}^{t-i}$, and concatenates the history set with the current crowd-flow state, $\vb{x}^t$, to form a tuple input, $\bigcup_{i=0}^{h} \vb{x}^{t-i}$, to the crowd-flow prediction model.
    
    We note that, for $1 \leq k \leq h$, the history set at time, $t$, is a union of a subset of the history set, $\bigcup_{i=1}^{l_h-k} \vb{x}^{(t-k)-i}$, and the crowd-flow state, $\vb{x}^{(t-k)}$, of the model input at time, $t-k$. An input is consistent, if and only if,
    \begin{equation}
        \forall k \in [1..h], \bigcup_{i=0}^{h-k} \vb{x}^{(t-k)-i} = \bigcup_{i=k}^{h} \vb{x}^{t-i}
        \label{eq:consistency_property}
    \end{equation}
    which leads to the consistency check mechanism that we develop later. Simply, the history set at any time, $t$, should be consistent with the crowd-flow states at the previous times as illustrated in Fig.~\ref{fig:consistency_illustration} with an example.
    
    \textbf{Remark.} For each $t$, a standard adversarial attack has to learn a new set of perturbations, $\bigcup_{i=0}^{h} \boldsymbol{\updelta}^{t-i}$, independent (and therefore, different) from the perturbations, $\bigcup_{i=0}^{h} \boldsymbol{\updelta}^{(t-k)-i}$, learned for some previous time, $t-k$. Formally, for standard adversarial attacks,
    \begin{equation}
        \forall k \in [1..h], 
        \bigcup_{i=k}^{h} \boldsymbol{\updelta}^{t-i} \neq \bigcup_{i=0}^{h-k} \boldsymbol{\updelta}^{t-k-j}
        \label{eq:attack_inconsistency}
    \end{equation}
    Stated simply, the adversarial perturbations (and hence, the adversarially perturbed inputs), generated by the standard adversarial attacks are \textit{inconsistent}, and therefore, can be readily detected by our \emph{CaV-detect} mechanism as formalized previously.
    
    \item \heading{Validity:} Consider a $4 \times 4$ grid shown in Fig.~\ref{fig:inflow_outflow_illustration}. For a grid point-$(0,1)$, (shaded green) the total number of devices entering into the grid point from its adjacent grid points (shaded yellow) is 2 (C2 and C3). As these devices must outflow from the adjacent grid points, the total outflow from the adjacent grid points must atleast be 2. More generally, we consider adjacency to be within the $2^{nd}$ neighborhood of the grid point as shown in Fig.~\ref{fig:inflow_outflow_illustration}. Given a specific grid point-$(p_1, p_2)$, let $A_n(p_1, p_2)$ denote a set of grid points adjacent to the grid point-$(p_1, p_2)$ in the $n^{th}$ neighborhood,
    \begin{equation}
        A_n(p_1, p_2) = \bigcup_{i=-n}^{n} \bigcup_{\substack{j=-n \\ |i| + |j| \neq 0}}^{n} (p_1-i, p_2-j)
        \label{eq:adjacent_gridpoints}
    \end{equation}
    where $n$ is the number of neighbors considered for adjacency. By definition, at any given time, $t$, the inflow to the grid point-$(p_1, p_2)$, is the total number of devices entering into that grid point from its adjacent grid points, $A_n(p_1, p_2)$. Therefore, the total outflow from $A_n(p_1, p_2)$ must be atleast equal to the total inflow to $(p_1, p_2)$. Let $\vb{x}^t_{in}(p_1, p_2)$ and $\vb{x}^t_{out}(p_1, p_2)$ respectively denote the inflow to and outflow from the grid point-$(p_1, p_2)$ at time, $t$. Any input to the crowd-flow prediction model is only \textit{valid}, if,
    \begin{subequations}
        \begin{align}
            \vb{x}^t_{in}(p_1, p_2) \leq \sum_{(p'_1, p'_2) \in A_n(p_1, p_2)} \vb{x}^t_{out}(p'_1, p'_2) \\
            \vb{x}^t_{out}(p_1, p_2) \leq \sum_{(p'_1, p'_2) \in A_n(p_1, p_2)} \vb{x}^t_{in}(p'_1, p'_2)
        \end{align}
        \label{eq:input_validity}
    \end{subequations}
    
    \textbf{Remark.} While generating adversarial perturbations, standard adversarial attacks formalized in eq-\eqref{eq:general_adversarial_goal} do not respect this relationship between the inflow and outflow matrices, and therefore, can be detected at run-time.
\end{enumerate}

\begin{figure}
    \centering
    \includegraphics[width=0.8\linewidth]{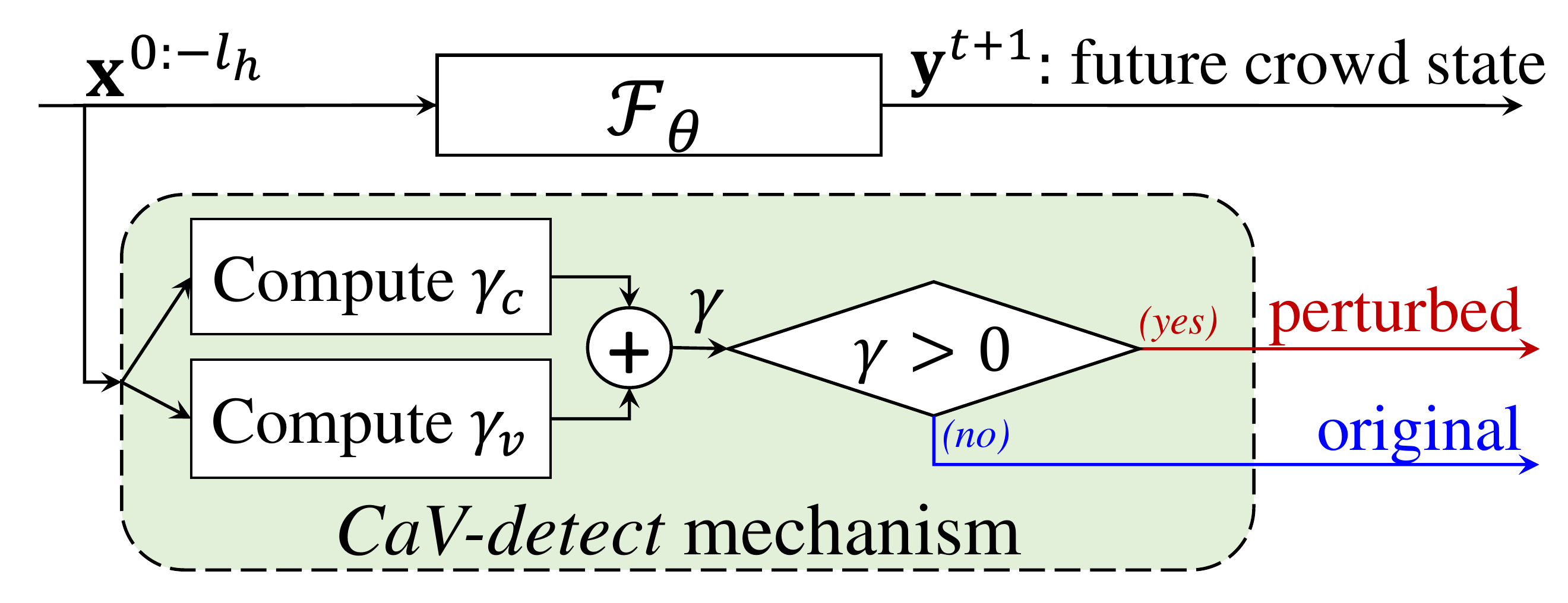}
    \caption{Illustrating our newly proposed \emph{CaV-detect} methodology integrated with the crowd-flow prediction model to detect adversarial inputs at run-time. For any input, \emph{CaV-detect} checks the consistency, $\gamma_c > 0$ and the validity $\gamma_v > 0$ of the input. The input is marked adversarial if any of the checks fail. \take{\emph{CaV-detect} does not require retraining the model and can be integrated with an off-the-shelf model.}}
    \label{fig:cavdetect_methodology}
\end{figure}
\subsection{CaV-detect: Consistency and Validity Check Mechanism to Detect Adversarially Perturbed Inputs}
Here we utilize the previously defined two properties of crowd-flow inputs to propose \emph{CaV-detect}, a novel input validation mechanism to detect adversarially perturbed inputs to the crowd-flow prediction models. To summarize, our \emph{CaV-detect} methodology comprises two main steps---\textit{consistency} check mechanism and \textit{validity} check mechanism. An input to the model is considered adversarially perturbed if it fails to satisfy either of the aforementioned checks. Step-by-step details of  \emph{CaV-detect} are given below.

\vspace{1mm}
\heading{Consistency check mechanism:} Let $\vb{x}^t$ denote the crowd-flow state at any given time, $t$. The input to the crowd-flow prediction model, $\mathcal{F}_{\theta^*}$, can then be denoted by $\vb{X}_h(t) = \bigcup_{i=0}^{h} \vb{x}^{t-i}$, where $h$ denotes the history length. Our consistency check mechanism works in two steps:

\begin{enumerate}
    \item Firstly, we keep all the model inputs, $\vb{x}^{(t-k)}$, received at the previous times, ${t-k}$, saved in the memory, $\forall k \in [1..h]$.
    
    \item Noting that the model inputs received at the previous times, ${t-k}$, reappear in the history set of the input received at the current time, $t$, we compute the difference between appropriately cropped model inputs at different times
    \begin{equation}
        \gamma_c = \sum_{k=1}^{h} \left| \bigcup_{i=k}^{h} \vb{x}^{t-i} - \bigcup_{i=0}^{h-k} \vb{x}^{(t-k)-i} \right| \geq 0
        \label{eq:consistency_indicator}
    \end{equation}
    where $\gamma_c$ is the inconsistency score---the closer $\gamma_c$ is to zero, the more consistent the input.
    
    \item The input is marked as adversarial if $\gamma_c > 0$.
\end{enumerate}

\vspace{1mm}
\heading{Validity check mechanism:} Let $\vb{x}^t = (\vb{x}_{in}^t, \vb{x}_{out}^t)$ denote the crowd-flow state at any given time, $t$, where $\vb{x}_{in}^t, \vb{x}_{out}^t \in \mathbb{R}^{l_1 \times l_2}$. Our validity check mechanism works in four steps described below.

\begin{enumerate}
    \item We first define a filter, $\vb{f} \in \mathbb{Z}^{5 \times 5}$, such that $\forall p_1,p_2 \in [0..4]$,
    \begin{equation}
    \vb{f}(p_1, p_2) = 
        \begin{cases}
            1, & p_1=2, p_2=2 \\
            0, & \text{otherwise}
        \end{cases}
    \end{equation}
    
    \item \label{p:inflow_outflow_validity_indicators} Secondly, we compute the inflow and outflow invalidity scores, denoted $\gamma_{v_i}$ and $\gamma_{v_o}$ respectively, by simultaneously analyzing both the inflow and outflow matrices in the input.
    \begin{align}
        \gamma_{v_i} = \vb{x}_{in}^{t-i} \circledast \vb{f} - \vb{x}_{out}^{t-i} \circledast (\vb{1} - \vb{f}) \leq \vb{0} \\
        \gamma_{v_o} = \vb{x}_{out}^{t-i} \circledast \vb{f} - \vb{x}_{in}^{t-i} \circledast (\vb{1} - \vb{f}) \leq \vb{0}
    \end{align}
    
    where $\circledast$ denotes a 2-D convolution operation.
    
    \item Finally, we compute the input invalidity score, $\gamma_v$, based on the inflow and outflow invalidity scores computed in step~\ref{p:inflow_outflow_validity_indicators}.
    \begin{equation}
        \gamma_v = \relu(\gamma_{v_i} + \gamma_{v_o})
        \label{eq:validity_indicator}
    \end{equation}
    where $\relu$ denotes the rectified linear unit function commonly used in DL literature.
    
    \item The input is marked as adversarial if $\gamma_v > 0$.
\end{enumerate}

Note that both the check mechanisms used by \emph{CaV-detect} are model agnostic. Therefore, \emph{CaV-detect} can be incorporated with the pre-trained crowd-flow prediction models of varying architectures without undermining their efficacy.

\begin{figure*}
    \centering
    \includegraphics[width=0.9\linewidth]{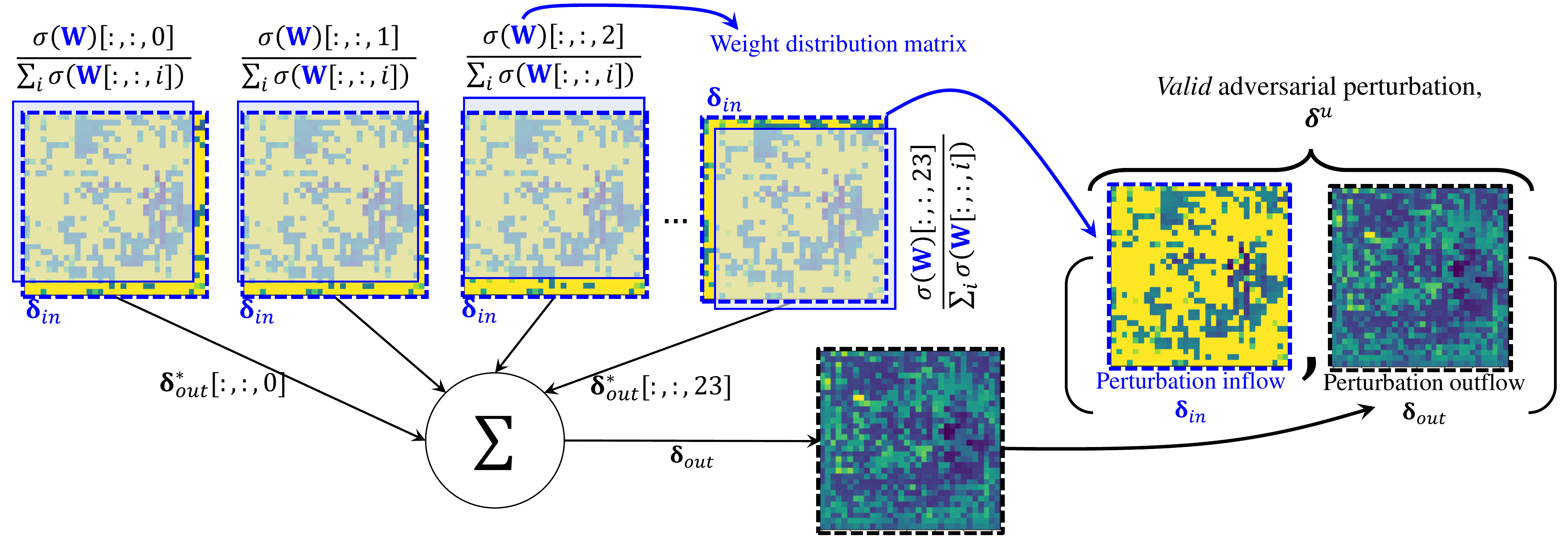}
    \caption{Illustrating the newly proposed \emph{CVPR} attack methodology of generating valid adversarial perturbations. Terms highlighted in blue denote the variables that are updated during attack to optimize the adversarial loss. Given a \textit{perturbation inflow matrix}, $\boldsymbol{\updelta}_{in}$, a set of \textit{distributed perturbation outflow matrices}, $\boldsymbol{\updelta}^*_{out}$ is computed by element-wise application of $\boldsymbol{\updelta}_{in}$ and the appropriately shifted normalized perturbation distribution matrices, $\vb{W}$. Finally, the \textit{perturbation outflow matrix}, $\boldsymbol{\updelta}_{out}$, is computed by adding all the slices of $\boldsymbol{\updelta}^*_{out}$.}
    \label{fig:validity_attack}
\end{figure*}
\subsection{CVPR-attack: Consistent Valid and Physically-Realizable Adversarial Attack against Crowd-flow Prediction Models}
In light of the previously formalized practical limitations of standard adversarial attacks, in this section, we propose \emph{CVPR} attack---a \textit{consistent}, \textit{valid} and \textit{physically realizable} adversarial attack. At any given time, $t$, we consider an input, $\vb{X}_h(t) = \bigcup_{i=0}^{h} \vb{x}^{t-i}$, to the model, $\mathcal{F}_\theta$. Our goal is to generate perturbations, $\boldsymbol{\Updelta}_h (t) = \bigcup_{i=0}^{h} \boldsymbol{\updelta}^{t-i}$, to the input in order to bring the model output closer to the adversarial target, $\vb{y}^{t+1}_{target}$.

\begin{enumerate}
    \item \heading{Consistency:} To ensure consistency in the perturbations, we leverage universal adversarial perturbations to regulate $\boldsymbol{\Updelta}_h (t)$, such that, $\forall i \in [0..h], \boldsymbol{\updelta}^{t-i} = \boldsymbol{\updelta}^u$.
    \begin{equation}
        \boldsymbol{\Updelta}_h (t) = \bigcup_{i=0}^{h} \boldsymbol{\updelta}^u = \bigcup_{i=0}^{h} (\boldsymbol{\updelta}^u_{in}, \boldsymbol{\updelta}^u_{out})
        \label{eq:cvpr_consistency}
    \end{equation}
    
    \item \heading{Validity:} To ensure validity, we introduce a novel mechanism to generate the \textit{perturbation outflow matrix}, $\boldsymbol{\updelta}^u_{out}$, given a \textit{perturbation inflow matrix}, $\boldsymbol{\updelta}^u_{in}$. More specifically, given $\boldsymbol{\updelta}^u_{in}$, a specific grid point-$(p_1, p_2)$, and a set of its adjacent grid points in the $n^{th}$ neighborhood, $A_n(p_1, p_2)$, we learn a perturbation distribution matrix, $\vb{W} \in \mathbb{R}^{l_1 \times l_2 \times (2n+1)^2-1}$, to first distribute the total perturbation inflow to $(p_1, p_2)$ among $A_n(p_1, p_2)$,
    \begin{equation}
        \boldsymbol{\updelta}^*_{out} = \boldsymbol{\updelta}^u_{in} \odot \frac{\sigma(\vb{W})}{\sum_i \sigma(\vb{W})[:,:,i]}
    \end{equation}
    
    where $\odot$ denotes the element-wise (Hadamard) multiplication $\boldsymbol{\updelta}^*_{out} \in \mathbb{R}^{l_1 \times l_2 \times (2n+1)^2-1}$ denotes a set of distributed perturbation outflow matrices for $\boldsymbol{\updelta}^*_{out}$ satisfying the validity condition of crowd-flow inputs. The total perturbation outflow for $(p_1, p_2)$ is then computed by accumulating relevant distributed outflows,
    \begin{equation}
        \boldsymbol{\updelta}^u_{out}(p_1, p_2) = \sum_{i=-n}^{n} \sum_{\substack{j=-n \\ |i|+|j| \neq 0}}^{n} \boldsymbol{\updelta}^*_{out}(p_1-i, p_2-j, k)
        \label{eq:cvpr_delta_out_generation}
    \end{equation}
    
    where $k$ is defined as,
    \begin{equation}
        k = 
        \begin{cases}
            (2n+1)(i+n)+j+n-1, & i>0, j>0 \\
            (2n+1)(i+n)+j+n, & \text{otherwise}
        \end{cases}
    \end{equation}
    
    In other words, $\boldsymbol{\updelta}_{out}$ is computed as a function, $f$, of $\boldsymbol{\updelta}_{in}$ and $\vb{W}$ as illustrated in Fig.~\ref{fig:validity_attack}. Eq-\eqref{eq:cvpr_consistency} can then be re-written as,
    \begin{equation}
        \boldsymbol{\Updelta}_h (t) = \bigcup_{i=0}^{h} (\boldsymbol{\updelta}^u_{in}, \boldsymbol{\updelta}^u_{out}) = \bigcup_{i=0}^{h} (\boldsymbol{\updelta}^u_{in}, f(\boldsymbol{\updelta}^u_{in}, \vb{W}))
        \label{eq:cvpr_validity}
    \end{equation}
    
    \item \heading{Physical Realizability:} In order to change the model output at any given time, $t$, the perturbations, $\boldsymbol{\Updelta}_h(t) = \bigcup_{i=0}^{h} \boldsymbol{\updelta}^{t-i}$, learned by the attacker are significantly different for different $i$. In practice, such attacks are only feasible under the digital attack setting. Realizing such attacks under the physical attack setting requires an attacker to precisely control the number of devices in each grid point, which is challenging because an attacker has to either repeatedly relocate the adversarial devices or have a sufficient number of adversarial devices repeatedly switched on and off to simulate $\boldsymbol{\Updelta}_h(t)$. Universal adversarial perturbation naturally addresses this by generating a single most effective perturbation for each time interval.
    
    Additionally, generating $\boldsymbol{\updelta}^u \in \mathcal{B}_\infty(\epsilon)$ ball only works under the digital attack setting. For physical attack setting, an attacker can only realize $\boldsymbol{\updelta}>0$ perturbations (for example, by physically adding a certain number of adversarial devices). Therefore, for physical attacks, we optimize the perturbations for $\mathcal{B}_\infty(0, \epsilon)$ bound.
\end{enumerate}

While generating the perturbations, $\boldsymbol{\updelta}^u$, we iteratively update $\boldsymbol{\updelta}_{in}$ and $\vb{W}$ to find the optimal perturbations. More specifically, we optimize the following loss function,

\begin{align}
    \mathcal{L}_{upa} \left(\vb{X}_h(t), (\boldsymbol{\updelta}^u_{in}, \vb{W}) \right) =
    \left| \mathcal{F} \left( \vb{X}_h(t)+ \bigcup_{i=0}^{h} \boldsymbol{\updelta}^u \right) - \vb{y}_{target}^{t+1} \right|, \nonumber\\
    \boldsymbol{\updelta}^u_{in}, \vb{W} = \argmax_{\boldsymbol{\updelta}^u_{in} \in \mathcal{B}(\epsilon), \vb{W}} -\mathcal{L}_{upa} \left(\vb{X}_h(t), (\boldsymbol{\updelta}^u_{in}, \vb{W}) \right)
\end{align}

where $\boldsymbol{\updelta}^u = (\boldsymbol{\updelta}^u_{in}, \boldsymbol{\updelta}^u_{out})$ denotes the universal adversarial perturbations that remain constant for all $\vb{x}^t \in \mathcal{D}$, $\epsilon$ is the maximum perturbation budget as discussed previously. For physical realizability (limitation 3), we repeatedly clip the negative values and project $\boldsymbol{\updelta}^u$ on $\mathcal{B}(\epsilon)$ ball while maximizing $\mathcal{L}_{adv}$ using gradient-descent. Details are given in Algirthm~\ref{alg:cvpr_attack}.




\begin{algorithm}[t]
    \caption{\emph{CVPR} attack Algorithm}
    \label{alg:cvpr_attack}
    \begin{algorithmic}[1]
    \Input
    \Statex $\vb{X}_h(t) \gets$ history of crowd-flow states
    \Statex $\mathcal{F} \gets$ trained model
    \Statex $\vb{y} \gets$ output crowd-flow state
    \Statex $N \gets \#$ of iterations
    \Statex $\epsilon \gets$ maximum perturbation budget
    \Output
    \Statex $\boldsymbol{\updelta}^u \gets (\boldsymbol{\updelta}^u_{in}, \boldsymbol{\updelta}^u_{out}) \gets$ universal adversarial perturbations
    \State Define $i \gets 1, \boldsymbol{\updelta}_{in} \gets \vb{0}, \vb{W} \gets \vb{-5}$
    \State Define $\eta \gets (5 \times \epsilon)/N$
    \Repeat
    \State $\boldsymbol{\updelta}^u \gets (\boldsymbol{\updelta}^u_{in}, f(\boldsymbol{\updelta}^u_{in}, \vb{W}))$
    \State $\mathcal{L}_{adv} \gets \left( \mathcal{F} \left( \vb{X}_h(t) + \bigcup_{i=0}^{h} \boldsymbol{\updelta}^u) \right) - \vb{y}_{target}^{t+1} \right)^2$
    \State $\boldsymbol{\updelta}^u_{in} \gets \boldsymbol{\updelta}^u_{in} - \eta \times \operatorname{sign}\left( \frac{\partial \mathcal{L}}{\partial \boldsymbol{\updelta}^u_{in}} \right)$
    \State $\vb{W} \gets \vb{W} - \eta \times \operatorname{sign}\left( \frac{\partial \mathcal{L}}{\partial \vb{W}} \right)$
    \State $\boldsymbol{\updelta}^u_{in} \gets \operatorname{clip}(-\epsilon, \epsilon)$ ------ ($l_\infty$ bound)
    \State $\boldsymbol{\updelta}^u_{out} \gets f(\boldsymbol{\updelta}^u_{in}, \vb{W})$
    \State $i \gets i+1$
    \Until $i \leq N$
    \end{algorithmic}
\end{algorithm}

\section{Experimental Setup}

\begin{figure}
    \centering
    \includegraphics[width=\linewidth]{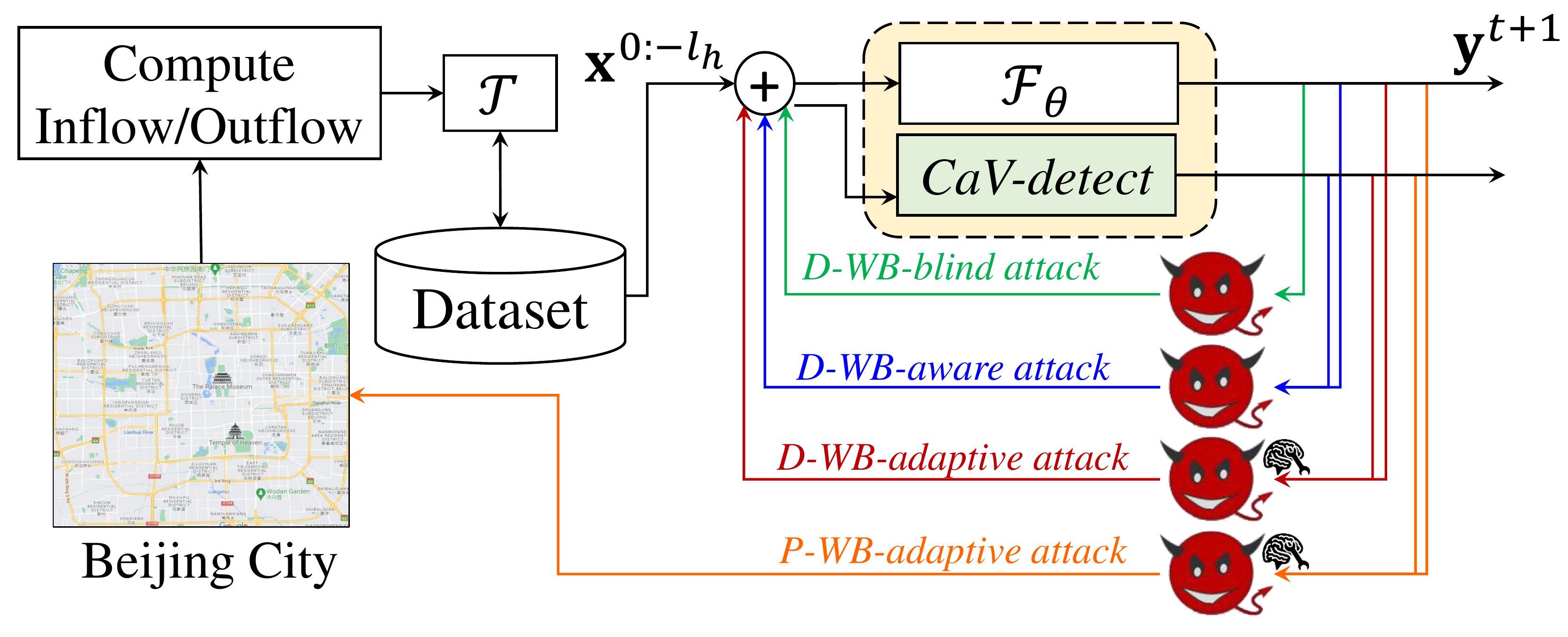}
    \caption{Illustration and comparison of different white-box threat models used in our experiments. D-WB denotes a digital attack setting under the white-box threat model. P-WB denotes a physical attack setting under white-box threat model. An adaptive attacker adapts the perturbation generation mechanism to fool \emph{CaV-detect}.}
    \label{fig:threat_models}
\end{figure}

\subsection{Threat Models}
For all the experiments in this paper, we assume a white-box threat model in which the attacker has complete knowledge of the crowd-flow prediction model architecture and its learned parameters, $\theta$. Further, we always assume a targeted attack scenario where the goal of an attacker is to perturb the input in order to realize a specific crowd-flow state, known as the target state denoted $\vb{y}_{target}^{t+1}$, at the output of the model. The goal of a white-box attacker is to learn the perturbations $\boldsymbol{\Updelta}_h(t) \in \mathcal{B}(\epsilon)$, that, when added to the inputs, produce the maximum relevance to an attacker's defined target state, $\vb{y}_{target}^{t+1}$.

\begin{equation}
    \mathcal{L}_{adv} \left(\vb{X}_h(t), \boldsymbol{\Updelta}_h(t) \right) =
    \left| \mathcal{F}(\vb{X}_h(t)+\boldsymbol{\Updelta}_h(t))
     - \vb{y}_{target}^{t+1}) \right|
     \label{eq:wb_goal}
\end{equation}

We experiment with three different white-box threat configurations as detailed below, and illustrated in Fig.~\ref{fig:threat_models}.

\vspace{1mm}
\heading{WB-blind Threat Model:} In this white-box threat model, the attacker is assumed to be unaware of the \emph{CaV-detect} mechanism deployed in the pipeline. The goal of WB-blind attacker is formalized below,

\begin{equation}
    \boldsymbol{\Updelta}_h(t) = \argmax_{\boldsymbol{\Updelta}_h(t) \in \mathcal{B}_{\infty}(\epsilon)} -\mathcal{L}_{adv} \left(\vb{X}_h(t), \boldsymbol{\Updelta}_h(t) \right)
    \label{eq:blind_attack}
\end{equation}

where $\mathcal{B}_{\infty}(\epsilon)$ is an $l_\infty$ ball, defined as, ${\boldsymbol{\updelta} \in \mathcal{B}_\infty(\epsilon) := -\epsilon \leq \boldsymbol{\updelta} \leq \epsilon}$. Eq-\eqref{eq:blind_attack} is iteratively optimized depending on the attack algorithm~\cite{szegedy2014intriguing}.

\vspace{1mm}
\heading{WB-aware Threat Model:} This white-box threat model assumes that the attacker is fully aware of \emph{CaV-detect} mechanism in the pipeline, and tries to evade the detection by \emph{CaV-detect} while simultaneously trying to produce the target state at the model output. Formally, we define a Lagrange function,

\begin{equation}
    \boldsymbol{\Updelta}_h(t) = \argmax_{\boldsymbol{\Updelta}_h(t) \in \mathcal{B}(\epsilon)} -\mathcal{L}_{adv} \left(\vb{X}_h(t), \boldsymbol{\Updelta}_h(t) \right) - \lambda \times (\gamma_c + \gamma_v)
    \label{eq:aware_attack}
\end{equation}

where $\lambda=10^{10}$ is the lagrange multiplier. Eq-\eqref{eq:aware_attack} is iteratively optimized depending on the attack algorithm.

\textbf{Note:} We conduct experiments under the WB-aware threat model and discover that a WB-aware attacker is unable to compute adversarial perturbations to cause any considerable change in the model output. We attribute this to the strict consistency and validity check mechanism, which leads to contradicting gradient updates while optimizing eq-\eqref{eq:aware_attack}. Therefore, we do not report the quantitative results in the paper.

\vspace{1mm}
\heading{WB-adaptive Threat Model:} Similar to WB-aware, this threat model also assumes an attacker who is fully aware of \emph{CaV-detect} mechanism in the pipeline and tries to evade the detection by \emph{CaV-detect} while significantly impacting the output towards the target state \textit{by adaptively modifying the attack algorithm}. More specifically, an adaptive attacker modifies the attack to make it easier for eq-\eqref{eq:aware_attack} to be optimized. In order to achieve this, our adaptive attacker leverages the algorithm of universal adversarial perturbations to naturally evade the consistency check mechanism. The adversarial loss function can then be defined as a Lagrange function below,

\begin{equation}
    \boldsymbol{\updelta}^{u} = \argmax_{\boldsymbol{\updelta}^u \in \mathcal{B}(\epsilon)} -\mathcal{L}_{adv} \left(\vb{X}_h(t), \bigcup_{i=0}^{h} \boldsymbol{\updelta}^{u} \right) - \lambda \times \gamma_v
    \label{eq:adaptive_attack}
\end{equation}

where we set $\lambda=10^{10}$. As previously, eq-\eqref{eq:adaptive_attack} is solved iteratively depending on the attack optimization algorithm.

\heading{Digital and Physical Settings:} In addition to the threat models mentioned above, we consider two different attack settings---digital and physical. The digital attack setting (D-WB) depicts a typical white-box scenario where an attacker is assumed to have the capability of hacking into the inference pipeline so that the attacker can directly perturb an input to the model. On the contrary, the physical attack setting (P-WB) depicts a more realistic white-box scenario where an attacker knows about the model's architecture and learned weights, but cannot directly perturb an input to the model. Therefore, an attacker has to instead physically add a certain number of devices, called adversarial devices, at specific grid points in order to realize adversarial perturbations. P-WB restricts an attacker by only allowing physical perturbations, which makes it more practical than D-WB. Nevertheless, due to its wide popularity in literature.


\subsection{Adversarial Attacks}
For D-WB settings, we evaluate three standard adversarial attacks---FGSM, i-FGSM, and PGD attacks---on our trained models. FGSM attack is simple, fast, and generates transferable adversarial perturbations, which makes it an ideal candidate for practical adversarial evaluation. On the other hand, PGD is among the strongest adversarial attacks found in literature against non-obfuscated models such as the ones we use in our evaluations. For P-WB settings, we compare the aforementioned attacks with our newly proposed \emph{CVPR} attack on different model architectures. 

\subsection{Evaluation Metrics}

\vspace{1mm}
\heading{Test Loss:} To evaluate a model, $\mathcal{F}$, on some test data, $\mathcal{D}_{test}$, we use a commonly used metric, the mean square error~(MSE), that captures the distance of the model output from the ground truth, $\vb{x}^{t+1}$, as defined below,

\begin{equation}
    \mathcal{L}(\mathcal{D}_{test}) = \frac{1}{|\mathcal{D}_{test}|} \sum_{\forall \vb{x}^t \in \mathcal{D}_{test}} (\mathcal{F}(\vb{X}_h(t)) - \vb{x}^{t+1})^2
    \label{eq:original_MSE}
\end{equation}

A smaller value of $\mathcal{L}(\mathcal{D}_{test})$ indicates a better learned model. 

\vspace{1mm}
\heading{Adversarial Loss:} For the adversarial evaluation, we let $\mathcal{D}^*_{test}$ denote the adversarially perturbed test data and compute an adversarial MSE, denoted $\mathcal{L}^*(\mathcal{D}^*_{test})$, as a measure of the model's robustness.

\begin{equation}
    \mathcal{L}^*(\mathcal{D}^*_{test}) = \frac{1}{|\mathcal{D}^*_{test}|} \sum_{\forall \vb{x}^t \in \mathcal{D}^*_{test}} (\mathcal{F}(\vb{X}_h(t)) - \vb{y}_{target}^{t+1})^2
    \label{eq:adv_MSE}
\end{equation}

where $\vb{y}_{target}^{t+1}$ denotes the targeted output desired by an attacker. A smaller value of $\mathcal{L}^*(\mathcal{D}^*_{test})$ indicates that the model is more \textit{robust} to adversarial perturbations and vice versa.

\vspace{1mm}
\heading{False Acceptance Rate (FAR):} To evaluate the efficacy of \textit{CaV-detector} to capture adversarial inputs, we use a widely used metric called FAR defined as the percentage of adversarial inputs marked unperturbed by the detector to the total number of adversarial inputs generated by the attacker. Additionally, we also use FAR to quantify the efficacy of adversarial attacks to evade the detection by \textit{CaV-detector}.

\subsection{Hyperparameters}
In this subsection, we report key hyperparameters that we analyze to understand the performance of the model under both, the standard and the adversarial scenarios.

\vspace{1mm}
\heading{Data:} While preparing the dataset, we use the history length, $h \in$ \{2, 5, 10, 15, 20\}.

\vspace{1mm}
\heading{Models:} We train different models based on the MLP and STResnet architectures by varying the number of hidden layers of each model. For the MLP model, we use the number of hidden layers in \{3, 5, 10\}. In the future, we denote the MLP model with $h$ hidden layers as \textit{mlp-h}. For STResnet model, we use the number of hidden residual blocks in \{1, 2, 3\} and denote them as \textit{STResnet-1, STResnet-2} and \textit{STResnet-3} respectively. For a TGCN model, we experiment with different dimensions of the hidden messages in \{1, 3, 5, 10\} (See \cite{zhao2019t} for the definition of hidden messages in the TGCN model) and study the effect of changing the number of neighbors, $d_A \in$ \{1, 3, 5, 10\} on the accuracy of TGCN models, where $d_A$ denotes the number of adjacent nodes of TGCN model assumed to be able to communicate with each other. For future reference, we denote \textit{tgcn-$(m, d_A)$} as a TGCN model with $m$ dimensional hidden messages and $d_A$ node connectivity.

\vspace{1mm}
\heading{Attacks:} For adversarial evaluation, we experiment with different maximum perturbations, $\epsilon \in$ \{0.01, 0.03, 0.05, 0.07, 0.1\} (for digital attack settings) and the maximum adversarial device budget, $b_d \in$ \{1000, 3000, 5000, 7000, 10000\} (for physical attack settings). We also analyze the effects of changing the maximum number of attack iterations, $N \in$ \{100, 250, 500, 750, 1000\} on the performance of the aforementioned model. For future references, we denote a specific attack setting as Attack-$(\epsilon, N)$.
For example, PGD-$(0.1, 1000)$ denotes a PGD attacker with the maximum allowed perturbation of $0.1$ and an iteration budge of $1000$.


\section{Results}
We first establish the baselines by reporting mean square loss over the original/unperturbed inputs. We then evaluate these models under the standard adversarial attacks and the newly proposed \emph{CVPR} attack. Finally, we show the efficacy of \emph{CVPR} attack over the standard adversarial attacks by comparing the number of adversarial devices required by each to achieve the adversarial goal.

\begin{figure*}
    \centering
    \begin{subfigure}{0.32\linewidth}
        \centering
        \includegraphics[width=1\linewidth, page=1]{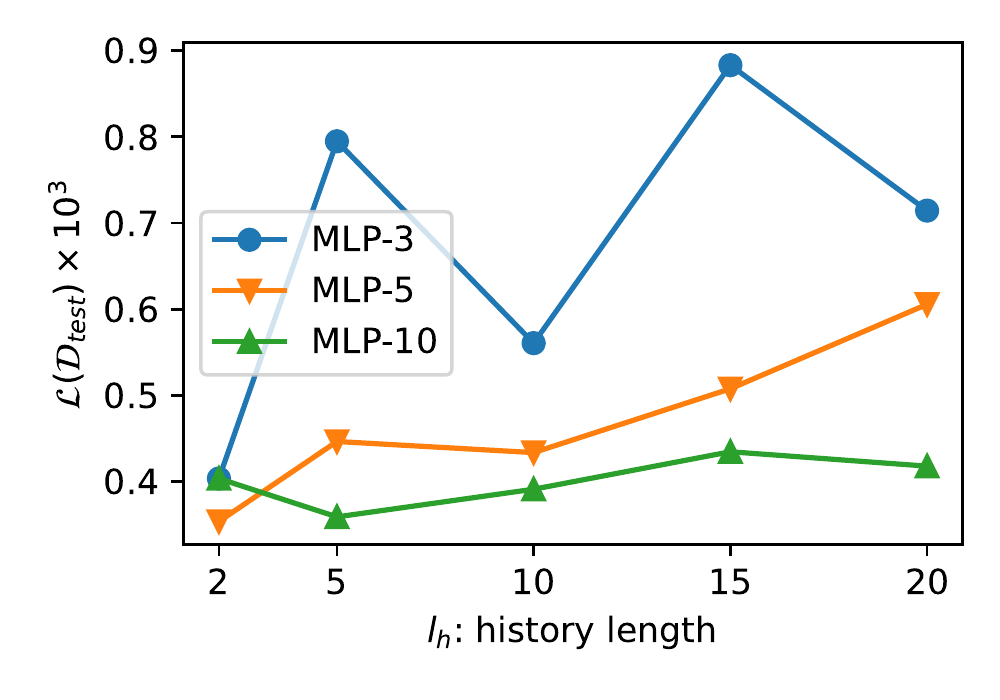}
        \caption{Multi-Layer Perceptron (MLP)}
    \end{subfigure}
    \begin{subfigure}{0.32\linewidth}
        \centering
        \includegraphics[width=1\linewidth, page=2]{Figures/clean_loss.pdf}
        \caption{Temporal Graph Convolution Network}
    \end{subfigure}
    \begin{subfigure}{0.32\linewidth}
        \centering
        \includegraphics[width=1\linewidth, page=3]{Figures/clean_loss.pdf}
        \caption{Spatio-Temporal ResNet (STResnet)}
    \end{subfigure}
    \caption{A comparison of \textbf{the model loss}, $\mathcal{L}(\mathcal{D}_{test})$ (eq-\eqref{eq:original_MSE}), over \textbf{the original/unperturbed test set}, $\mathcal{D}_{test}$, for different model complexities as the predefined history length, $h$, is increased. \settings{Dataset is TaxiBJ-16}. \take{No strict relationship between the model complexity and its performance over $\mathcal{D}_{test}$ is observed. When the input history length is increased, the $\mathcal{L}(\mathcal{D}_{test})$ increases in a slight manner, indicating a decrease in model performance. Of the three architectures, STResnet performs best.}}
    \label{fig:lh_original_loss}
\end{figure*}

\subsection{Performance of prediction models}\label{sec:results_performance}
Fig.~\ref{fig:lh_original_loss}(a, b and c) compare $\mathcal{L}(\mathcal{D}_{test})$ over unperturbed test inputs of MLP, TGCN and STResnet models (with different levels of complexity) trained on TaxiBJ-16 dataset with different history lengths.
We do not observe any strict relationship between the complexity of a model and its performance over unperturbed test inputs for the hyperparameters chosen in this experiment. We also note that STResnet models generally perform better than MLP and TGCN models which can be attributed to their ability to capture spatio-temporal relationships in the data due to the priors encoded in their architecture. Although TGCN models also capture spatio-temporal patterns in data, they have far fewer parameters as compared to STResnet models.

In Fig.~\ref{fig:lh_original_loss}, we observe that when the input history length is increased, the $\mathcal{L}(\mathcal{D}_{test})$ of the models generally increases, though very slightly. The trend is observed for all three model architectures considered in this paper with occasional exceptions. We hypothesize that a greater history length increases the input information to the model, which may lead to mutually contradicting gradient updates during training causing the resulting model to underfit. For relatively simpler models that are already vulnerable to underfitting, the increase in $\mathcal{L}(\mathcal{D}_{test})$ with the increase in $h$ is more significant, which further validates our hypothesis.


\begin{figure*}
    \centering
    \begin{subfigure}{0.32\linewidth}
        \centering
        \includegraphics[width=1\linewidth, page=1]{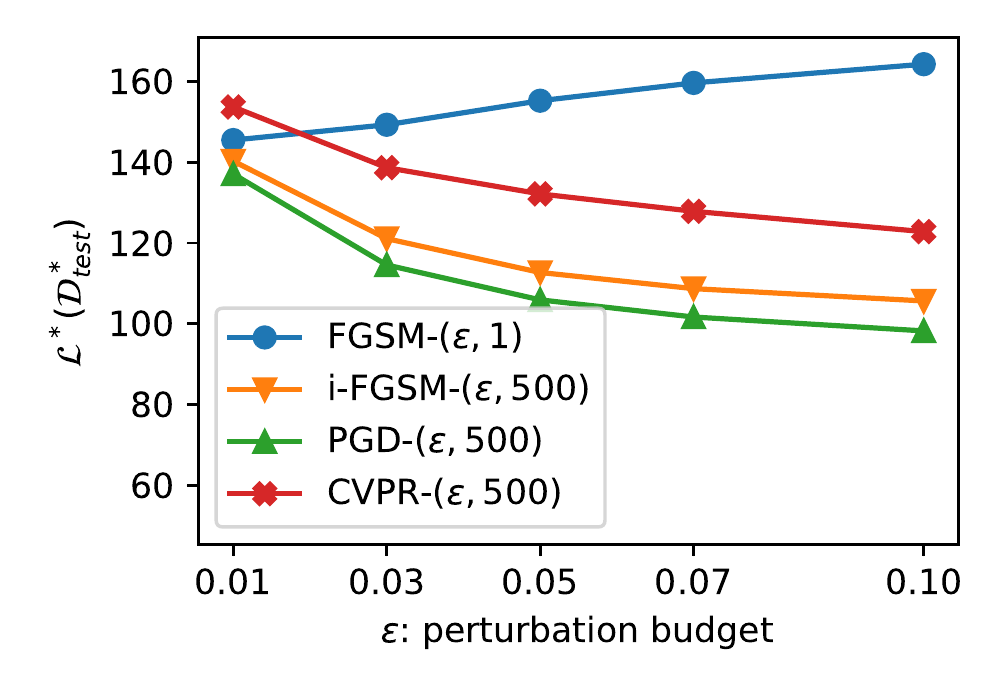}
        \caption{MLP-5 model}
    \end{subfigure}
    \begin{subfigure}{0.32\linewidth}
        \centering
        \includegraphics[width=1\linewidth, page=2]{Figures/non_adaptive_attack_results_upa_loss.pdf}
        \caption{TGCN-(5,5) model}
    \end{subfigure}
    \begin{subfigure}{0.32\linewidth}
        \centering
        \includegraphics[width=1\linewidth, page=3]{Figures/non_adaptive_attack_results_upa_loss.pdf}
        \caption{STResnet-2 model}
    \end{subfigure}
    \caption{Comparing \textbf{the adversarial loss}, $\mathcal{L}^*(\mathcal{D}^*_{test})$ (eq-\eqref{eq:adv_MSE}), over \textbf{the perturbed dataset}, $\mathcal{D}^*_{test}$, by different attacks for different model architectures as $\epsilon$ is increased assuming \textbf{a D-WB-blind attacker}. \settings{Dataset is TaxiBJ-16; $h$ is 5.}. \take{Deep crowd-flow prediction models are vulnerable to adversarial attacks. Increasing $\epsilon$ makes the attack stronger. TGCN-(5,5) model is the most robust of the considered architectures.}}
    \label{fig:eps_advloss_nonadaptive}
\end{figure*}

\subsection{D-WB-blind Adversarial Attacks}
In this experiment, we assume that the attacker is not aware of the \emph{CaV-detect} mechanism, and hence the attacker assumes a vanilla model when attacking. Fig.~\ref{fig:eps_advloss_nonadaptive}(a-c) summarizes our results of four adversarial attacks---FGSM-$(\epsilon, 1)$, i-FGSM-$(\epsilon, 500)$, PGD-$(\epsilon, 500)$ and \emph{CVPR} attack-$(\epsilon, 500)$---on the crowd-flow prediction models of different architectures---MLP-5, TGCN-(5,5), and STResnet-2---for different perturbation budgets, $\epsilon \in \{0.01, 0.03, 0.05, 0.07, 0.1\}$. Overall, we note that deep crowd-flow prediction models, like other deep learning models, are significantly vulnerable to adversarial attacks as illustrated by considerably smaller values of $\mathcal{L}^*(\mathcal{D}^*_{test})$ for $\epsilon>0$ compared to those for $\epsilon=0$.

As evident in the figure, increasing $\epsilon$ makes the attack stronger which is illustrated by a corresponding decrease in the value of $\mathcal{L}^*(\mathcal{D}^*_{test})$. These results are consistent with our intuitions, as a greater $\epsilon$ lets an attacker have a greater influence over the inputs of the model which in turn control the output. Overall, the PGD attack appears to be the strongest of all attacks, while our \emph{CVPR} attack appears the weakest. However, as we will see later, the adversarial perturbations generated by the three standard attacks are 100\% detectable by our \emph{CaV-detect} mechanism, while the perturbations generated by the \emph{CVPR} attack remain undetected.

We note that, as compared to MLP-5 and TGCN-(5,5), STResnet-2 model exhibits smaller values of $\mathcal{L}^*(\mathcal{D}^*_{test})$. This suggests that STResnet-2 model is relatively more vulnerable to adversarial perturbations, which appears surprising, as we have observed that STResnet-2 model gives comparatively better performance on the unperturbed dataset, $\mathcal{D}_{test}$, as compared to other architectures. These observations hint at the possibility of an accuracy-robustness tradeoff in the crowd-flow prediction models, as has been commonly observed in other DL models~\cite{ali2021all, wu2021wider}.

Unlike other architectures, for the MLP-5 model, the adversarial loss, $\mathcal{L}^*(\mathcal{D}^*_{test})$, slightly increases as the maximum perturbation budget, $\epsilon$, is increased. This is because FGSM is a single-shot attack, and the gradients computed by the attacker only estimate the loss surface within a limited range of input perturbations. Larger perturbations render these gradients imprecise, thus, degrading the efficacy of the attack. On the other hand, an i-FGSM attacker iteratively computes these gradients after small perturbation steps, which significantly increases the strength of the attack.

\begin{figure*}
    \centering
    \begin{subfigure}{0.32\linewidth}
        \centering
        \includegraphics[width=1\linewidth, page=1]{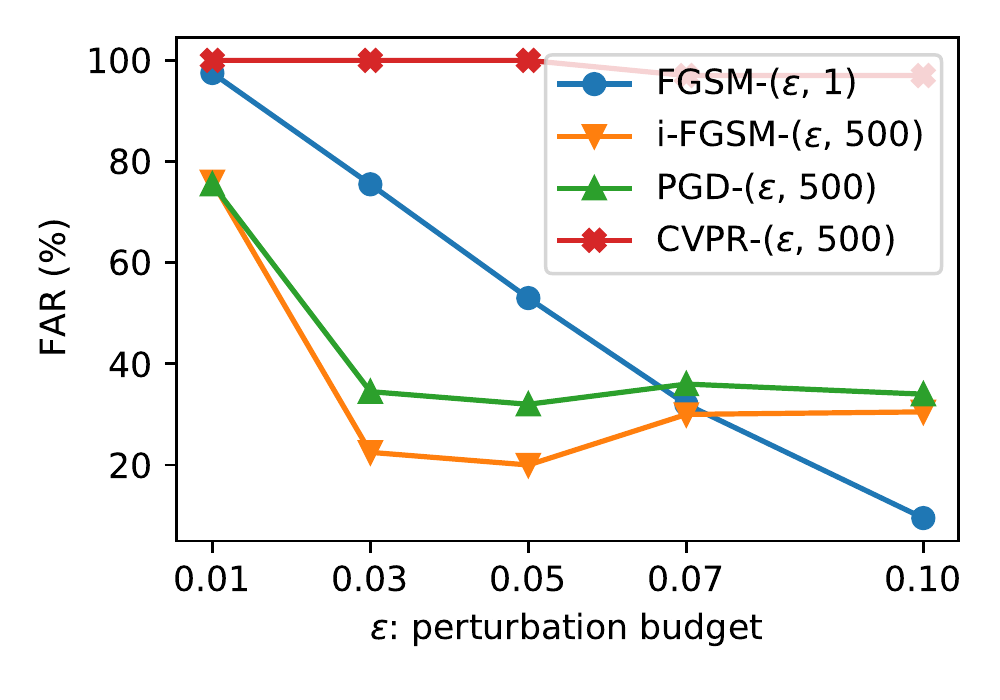}
        \caption{MLP-5 model}
    \end{subfigure}
    \begin{subfigure}{0.32\linewidth}
        \centering
        \includegraphics[width=1\linewidth, page=2]{Figures/non_adaptive_validity.pdf}
        \caption{TGCN-(5,5) model}
    \end{subfigure}
    \begin{subfigure}{0.32\linewidth}
        \centering
        \includegraphics[width=1\linewidth, page=3]{Figures/non_adaptive_validity.pdf}
        \caption{STResnet-2 model}
    \end{subfigure}
    \caption{False acceptance rate (FAR) of \emph{CaV-detect} mechanism against the perturbed inputs, $\mathcal{D}^*_{test}$, generated by \textbf{a D-WB-blind attacker}. \settings{Dataset is TaxiBJ-16. $h$ is 5}. \take{The adversarial perturbations become increasingly invalid as $\epsilon$ increases. FAR of the consistency check mechanism is always 0\%, so we only report FAR of the validity-check mechanism.}}
    \label{fig:cavdetect_blind}
\end{figure*}

\vspace{1mm}
\heading{Detecting D-WB-blind adversarial perturbations:} In this experiment, we evaluate the efficacy of consistency and validity properties to detect adversarial perturbations. We assume a D-WB-blind attacker as illustrated in Fig.~\ref{fig:threat_models}. We process the adversarial perturbations generated by the standard adversarial attacks through our novel \emph{CaV-detect} mechanism and report the results for MLP-5, TGCN-(5,5), and STResnet-2. In summary, with the False Rejection Rate (FRR) set to $\leq$0.5\%, our \emph{CaV-detect} mechanism shows a False Acceptance Rate (FAR) of 0\% against standard adversarial attacks and FAR of $>$99.7\% against our proposed \emph{CVPR} attack.

In Fig.~\ref{fig:cavdetect_blind}, we only report the FAR of the validity check mechanism. This is because in our experiment (D-WB-blind attacker), the FAR of the consistency check mechanism is always 0\% against standard adversarial attacks and 100\% against \emph{CVPR} attack. Consequently, the overall FAR of \emph{CaV-detect} is 0\% for standard adversarial attacks and equal to the FAR of validity check mechanism for \emph{CVPR} attack.

In Fig.~\ref{fig:cavdetect_blind}, we observe that as $\epsilon$ increases, the adversarial perturbations become increasingly invalid. This is not surprising, because a crowd-flow input, $\vb{X}_h(t)$, is initially valid, and introducing invalid perturbations of larger magnitude more significantly affects the validity of the perturbed inputs. For the MLP-5 model, we observe that the generated adversarial perturbations are relatively more valid as compared to TGCN-(5,5) and STResnet-2 models. We conjecture that because of its relatively simpler architecture, the features learned by the MLP-5 model are mostly linear, which leads to the linear gradients w.r.t. the model inputs. By definition in eq-\eqref{eq:input_validity}, if $\vb{X}_h(t)$ is valid, its linear multiple is also valid.

\begin{figure*}
    \centering
    \begin{subfigure}{1\linewidth}
        \centering
        \includegraphics[width=0.32\linewidth, page=1]{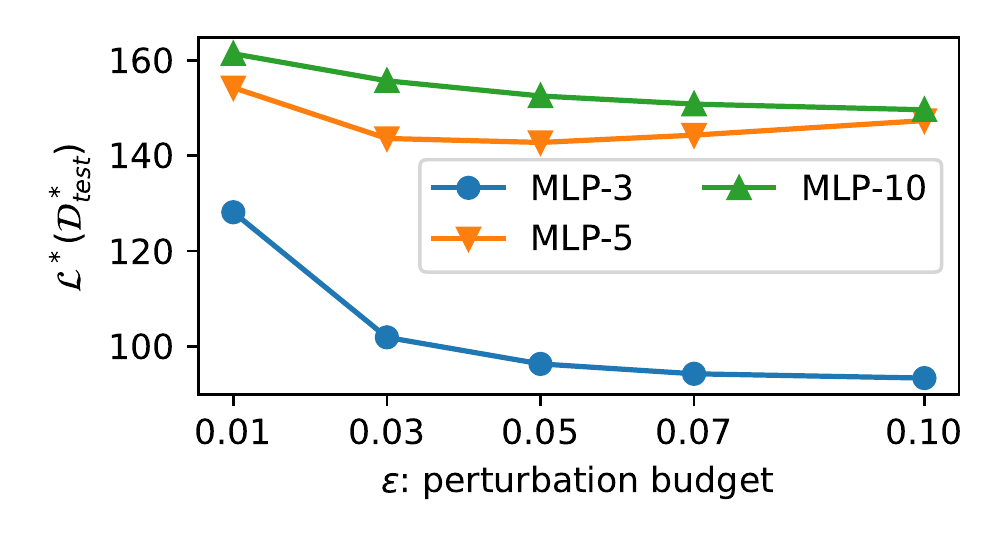}
        \includegraphics[width=0.32\linewidth, page=2]{Figures/attack_results_fgsm_loss_adaptive.pdf}
        \includegraphics[width=0.32\linewidth, page=3]{Figures/attack_results_fgsm_loss_adaptive.pdf}
        \caption{Comparing the adversarial loss, $\mathcal{L}^*(\mathcal{D}^*_{test})$, over the perturbed dataset, $\mathcal{D}^*_{test}$, by FGSM-($\epsilon$, 1) attack for different model architectures.}
    \end{subfigure}
    
    \begin{subfigure}{1\linewidth}
        \centering
        \includegraphics[width=0.32\linewidth, page=1]{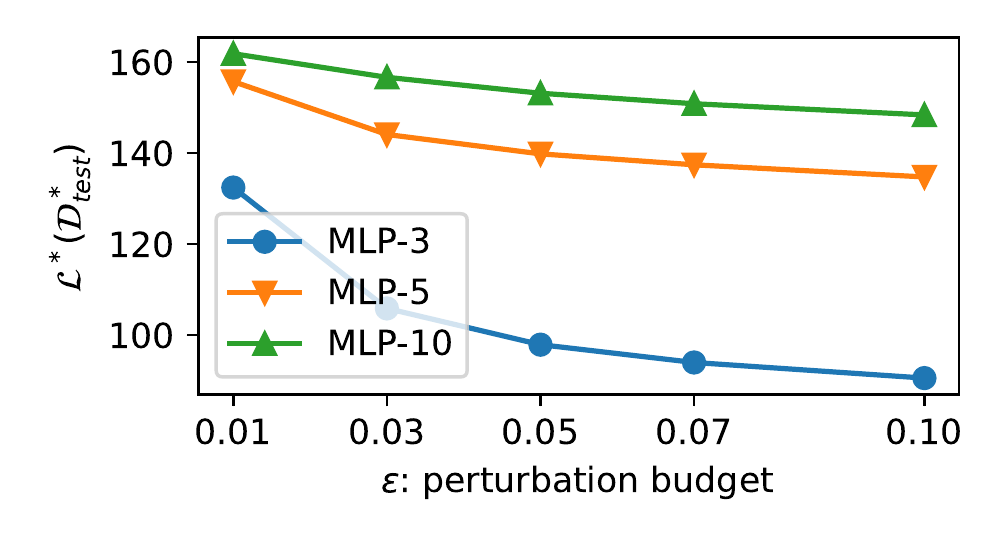}
        \includegraphics[width=0.32\linewidth, page=2]{Figures/attack_results_ifgsm_loss_adaptive.pdf}
        \includegraphics[width=0.32\linewidth, page=3]{Figures/attack_results_ifgsm_loss_adaptive.pdf}
        \caption{Comparing the adversarial loss, $\mathcal{L}^*(\mathcal{D}^*_{test})$, over the perturbed dataset, $\mathcal{D}^*_{test}$, by iFGSM-($\epsilon$, 500) attack for different model architectures.}
    \end{subfigure}
    
    \begin{subfigure}{1\linewidth}
        \centering
        \includegraphics[width=0.32\linewidth, page=1]{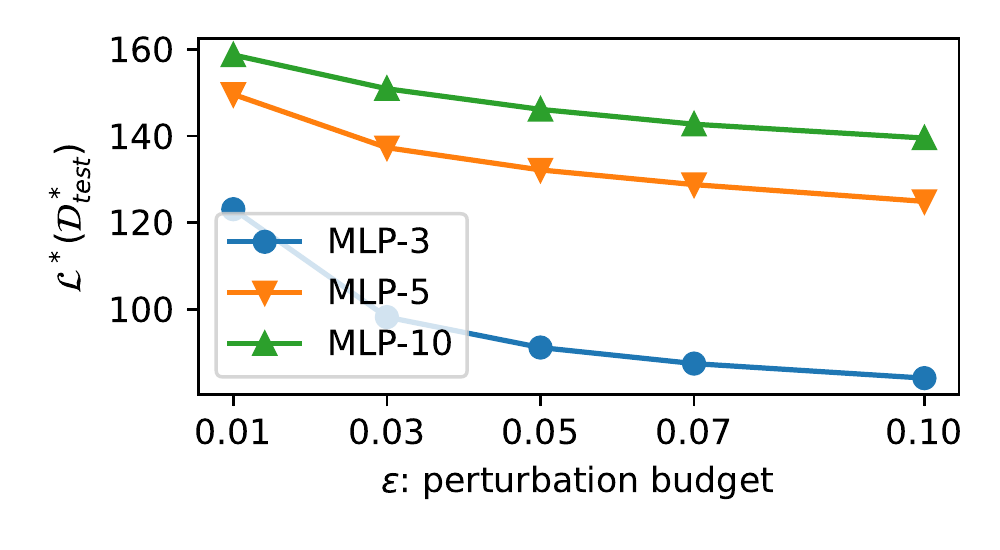}
        \includegraphics[width=0.32\linewidth, page=2]{Figures/attack_results_pgd_loss_adaptive.pdf}
        \includegraphics[width=0.32\linewidth, page=3]{Figures/attack_results_pgd_loss_adaptive.pdf}
        \caption{Comparing the adversarial loss, $\mathcal{L}^*(\mathcal{D}^*_{test})$, over the perturbed dataset, $\mathcal{D}^*_{test}$, by PGD-($\epsilon$, 500) attack for different model architectures.}
    \end{subfigure}
    
    \begin{subfigure}{1\linewidth}
        \centering
        \includegraphics[width=0.32\linewidth, page=1]{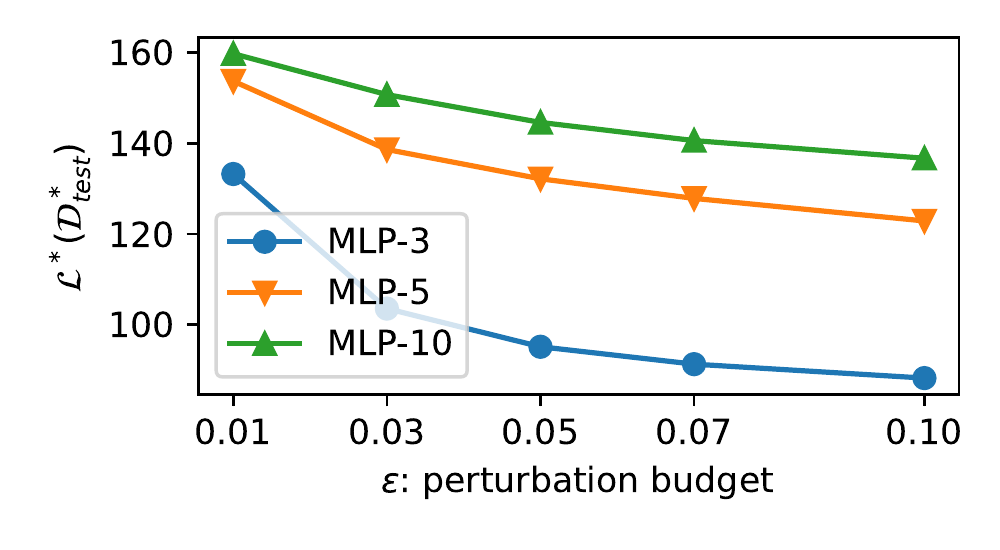}
        \includegraphics[width=0.32\linewidth, page=2]{Figures/attack_results_upa_loss_adaptive.pdf}
        \includegraphics[width=0.32\linewidth, page=3]{Figures/attack_results_upa_loss_adaptive.pdf}
        \caption{Comparing the adversarial loss, $\mathcal{L}^*(\mathcal{D}^*_{test})$, over the perturbed dataset, $\mathcal{D}^*_{test}$, by CVPR-($\epsilon$, 500) attack for different model architectures.}
    \end{subfigure}
    \caption{Comparing \textbf{the adversarial loss}, $\mathcal{L}^*(\mathcal{D}^*_{test})$, over  assuming \textbf{a D-WB-blind attacker}, $\mathcal{D}^*_{test}$, by different attack algorithms (a)-(d) for different model architectures as $\epsilon$ is increased assuming \textbf{a D-WB-adaptive attacker}. \settings{Dataset is TaxiBJ-16; $l_h$ is 5}. \take{Of the three architectures assumed in the paper, the TGCN-(5,5) model shows the greatest adaptive adversarial robustness against different attacks followed by the STResnet-2 model.}.}
    \label{fig:epsilon_adv_loss_adaptive}
\end{figure*}
\subsection{D-WB-adaptive Adversarial Attacks}
In this experiment, we assume that the attacker is aware of the \emph{CaV-detect} mechanism, and is able to adaptively reformulate the attack methodology to single-handedly fool both the crowd-flow prediction models and \emph{CaV-detect} mechanism. Fig.~\ref{fig:eps_advloss_nonadaptive}(a-d) summarizes our results of four different adversarial attacks---FGSM-$(\epsilon, 1)$, i-FGSM-$(\epsilon, 500)$, PGD-$(\epsilon, 500)$ and \emph{CVPR} attack-$(\epsilon, 500)$---on the crowd-flow prediction models assumed  in this paper for different perturbation budgets, $\epsilon \in \{0.01, 0.03, 0.05, 0.07, 0.1\}$. 

As observed previously, increasing $\epsilon$ increases the strength of the attack, thus more significantly affecting the output of the model as illustrated by a decreased $\mathcal{L}^*(\mathcal{D}^*_{test})$ values. Contrary to our previous observation (where the STResnet-2 model was least robust), we note that against adaptive attacks, the robustness of STResnet models is on par with or better than MLP models. Of the three architectures assumed in the paper, the TGCN model shows the greatest adaptive adversarial robustness.

For the MLP architecture, we note that a deeper MLP-10 model is more robust as compared to MLP-3 and MLP-5 models. These observations are coherent with a previous study~\cite{ali2021all}, which shows that increasing the model complexity slightly increases the model robustness. We observe that all the standard attacks considered in this paper give a comparable adversarial performance, which is counter-intuitive, as i-FGSM and PGD are iterative attacks, and are generally considered stronger than the FGSM attack. We attribute this to the inherent simplicity of MLP architecture, which fails to learn robust features from the input data, and thus can be equally manipulated by relatively simpler attacks.

For the TGCN architecture, we do not observe any strictly definitive effect of increasing a model's complexity on its adversarial robustness. For example, TGCN-(5,5), which communicates 5-dimensional hidden messages, is typically more robust as compared to the simpler TGCN-(1,5) and more complex TGCN-(10,5), which respectively communicate 1-dimensional 10-dimensional hidden messages. Overall our proposed \emph{CVPR} attack performs significantly better than the adaptive PGD attack, which in turn outperforms the adaptive i-FGSM attack by a large margin. i-FGSM and FGSM attacks show comparable performance with i-FGSM being slightly better than the FGSM in some cases.

For the STResnet architecture, STResnet-1 shows a considerably greater adversarial robustness followed by STResnet-2 which in turn leads STResnet-3 by a notable margin, which can be attributed to the increased adversarial vulnerability of latent DNN layers~\cite{kumari2019harnessing}. Yet again, we observe that adaptive PGD and \emph{CVPR} attacks perform significantly better than adaptive FGSM and adaptive i-FGSM attacks, specifically notable for $\epsilon=0.1$. For example, for STResnet-2 model, $\mathcal{L}^*(\mathcal{D}^*_{test}) \approx 100$ for adaptive PGD and \emph{CVPR} attacks versus $\max \mathcal{L}^*(\mathcal{D}^*_{test}) \approx 145$ for adaptive FGSM and adaptive i-FGSM attacks.

\begin{figure*}
    \centering
    \begin{subfigure}{0.32\linewidth}
        \centering
        \includegraphics[width=1\linewidth, page=1]{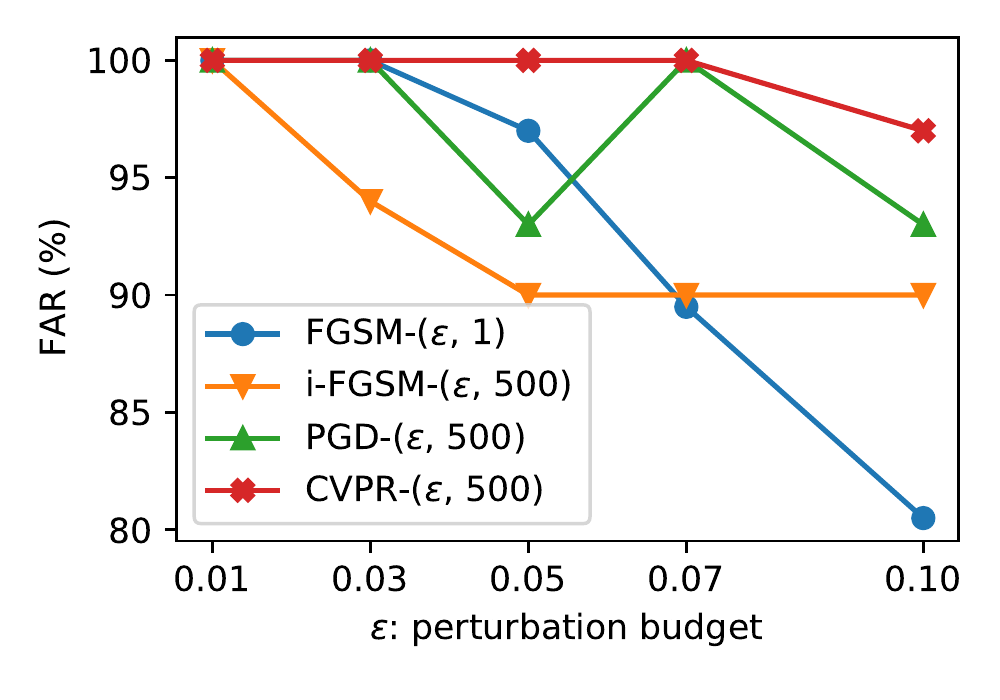}
        \caption{MLP-5}
    \end{subfigure}
    \begin{subfigure}{0.32\linewidth}
        \centering
        \includegraphics[width=1\linewidth, page=2]{Figures/adaptive_validity.pdf}
        \caption{TGCN-(5,5)}
    \end{subfigure}
    \begin{subfigure}{0.32\linewidth}
        \centering
        \includegraphics[width=1\linewidth, page=3]{Figures/adaptive_validity.pdf}
        \caption{STResnet-2}
    \end{subfigure}
    \caption{False acceptance rate (FAR) of \emph{CaV-detect} mechanism against the perturbed inputs, $\mathcal{D}^*_{test}$, generated by \textbf{a D-WB-adaptive attacker}. \settings{Dataset is TaxiBJ-16. $h$ is 5}. \take{The adversarial perturbations become increasingly invalid as $\epsilon$ increases. FAR of the consistency check mechanism is always 100\%, so we only report FAR of the validity-check mechanism.}}
    \label{fig:cavdetect_aware}
\end{figure*}

\vspace{1mm}
\heading{Detecting D-WB-adaptive adversarial perturbations:} In this experiment, we evaluate if the adversarial inputs perturbed adaptively can be detected based on the consistency and validity properties formalized previously. We assume a D-WB-adaptive attacker as illustrated in Fig.~\ref{fig:threat_models}. We process the adversarial perturbations generated by the adaptive adversarial attacks through \emph{CaV-detect} and report the results for MLP-5, TGCN-(5,5) and STResnet-2 models. In summary, with the False Rejection Rate (FRR) set to $\leq$0.5\%, our \emph{CaV-detect} mechanism shows considerably higher FAR ($\approx$80\%-100\%) against adaptive adversarial attacks as compared to that for non-adaptive standard attacks ($\approx$0\%) and FAR of $>$99.7\% against our proposed \emph{CVPR} attack.

In Fig.~\ref{fig:cavdetect_aware}, we only report the FAR of the validity check mechanism under adaptive attack settings. This is because in our experiment (D-WB-adaptive attacker), the FAR of the consistency check mechanism is always 100\% against all adaptive adversarial attacks due to the universal adversarial perturbations. Consequently, the overall FAR of \emph{CaV-detect} is equal to the FAR of the validity check mechanism.

As previously observed, increasing $\epsilon$ considerably decreases FAR, showing that the adversarial perturbations become increasingly invalid. Contrary to the \textit{CaV-Detect}-blind attacks, the adversarial inputs generated by D-WB-adaptive attacks can evade the detection mechanism with around 80\% FAR for standard adversarial attacks. We specifically attribute this to the newly proposed adaptive modifications to the standard attacks---\textit{universalizing} the adversarial perturbations and Lagrange optimization.

However, despite its D-WB-adaptive algorithm, we note that FGSM attack fails to perform well against \emph{CaV-detect} as illustrated by significantly reduced FARs, particularly notable for TGCN-(5,5) and STResnet-2 models where FAR drops to 0\% when $\epsilon=0.1$. Again, we attribute this to imprecise gradients for large perturbations due to FGSM being a single-shot attack, which significantly degrades the efficacy of the attack.

\section{Discussions}

\begin{figure*}
    \centering
    \begin{subfigure}{0.32\linewidth}
        \centering
        \includegraphics[width=1\linewidth]{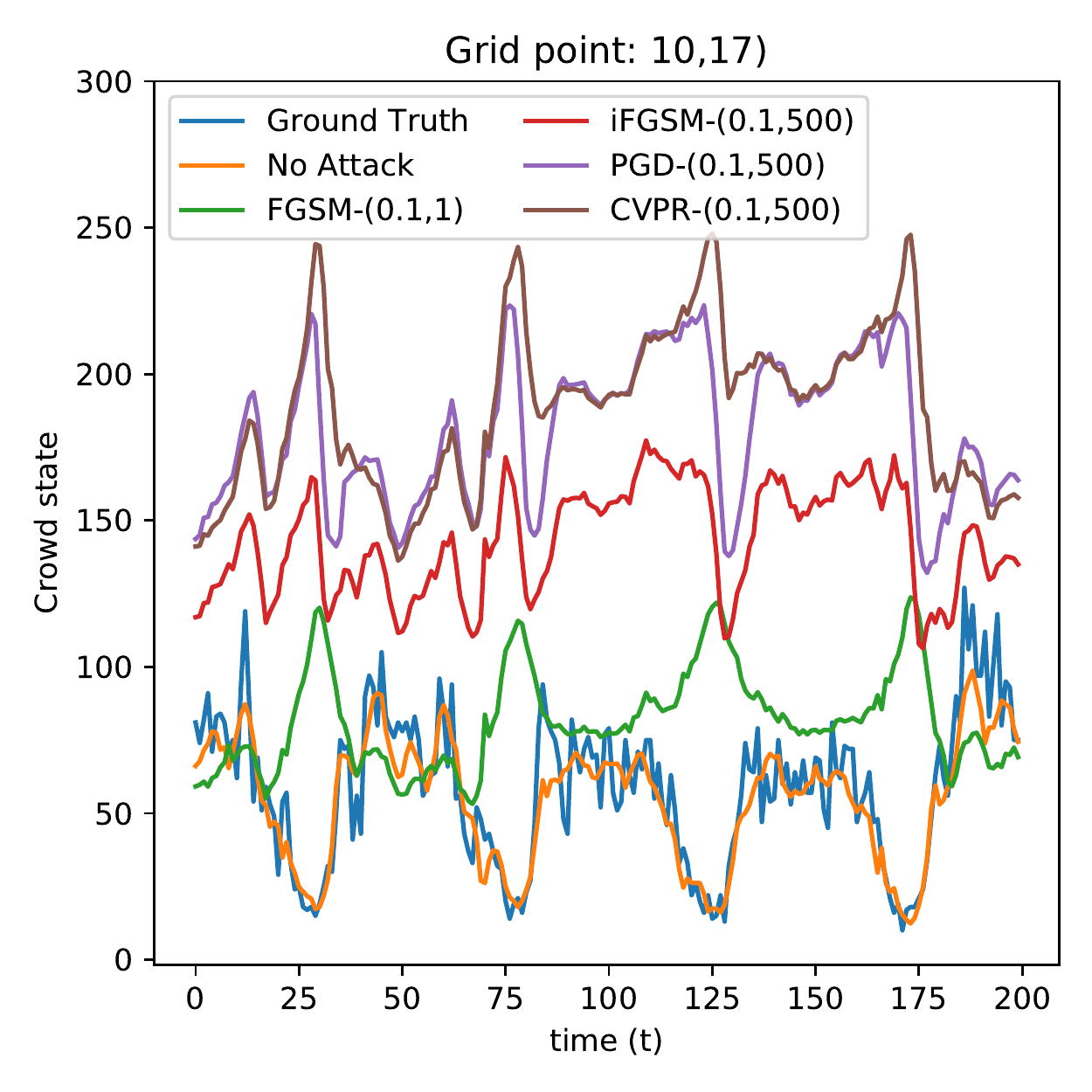}
        \caption{MLP-5 model}
    \end{subfigure}
    \begin{subfigure}{0.32\linewidth}
        \centering
        \includegraphics[width=1\linewidth]{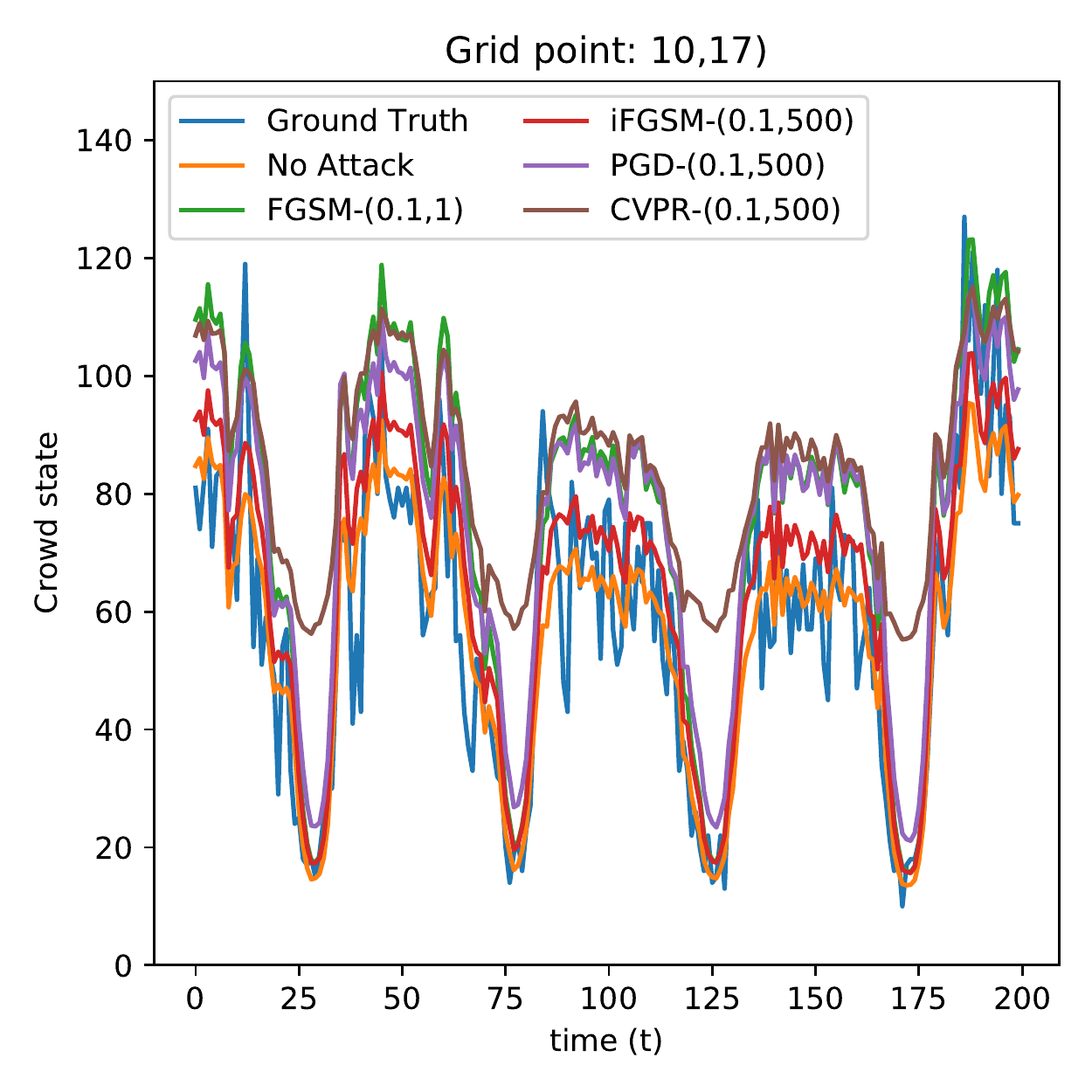}
        \caption{TGCN-(5,5) model}
    \end{subfigure}
    \begin{subfigure}{0.32\linewidth}
        \centering
        \includegraphics[width=1\linewidth]{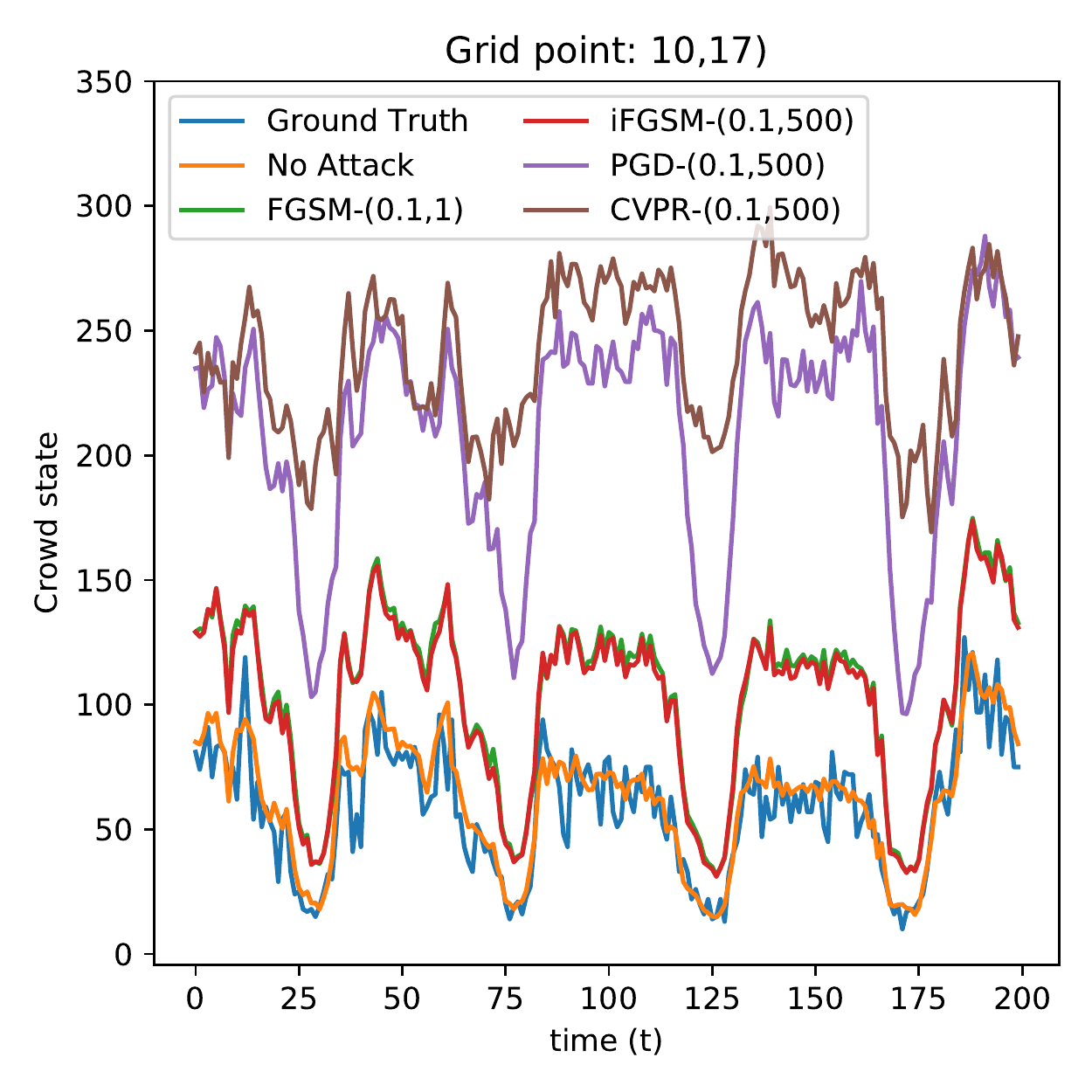}
        \caption{STResnet-2 model}
    \end{subfigure}
    \caption{Visualizing the predicted inflow states of the crowd-flow prediction models of different architectures with the actual inflow states (recorded in the future). ``No Attack'' denotes the predicted inflow states for the original/unperturbed inputs assuming \textbf{a D-WB-adaptive attacker}. \settings{Dataset is TaxiBJ-16; $h$ is 5; $\epsilon$ is 0.1}. \take{\emph{CVPR} attack outperforms the other attacks. TGCN-(5,5) model is more robust to consistent and valid adversarial attacks than the other two models.}}
    \label{fig:visualizing_outputs}
\end{figure*}

\subsection{Visualizing Crowd-flow Predictions}
Fig.~\ref{fig:visualizing_outputs} compares the predicted inflow states of different crowd-flow prediction models for different future times with the actual inflow states (ground truths recorded in the future) for both the original and the perturbed inputs generated by different adaptive attacks, where the goal of the attacks is to increase the predicted inflow state as much as possible while keeping the perturbations $\boldsymbol{\updelta} \in \mathcal{B}_\infty(\epsilon)$. The qualitative analysis shows that the \emph{CVPR} attack outperforms other attacks, as the predicted inflow state for \emph{CVPR}-attacked inputs typically exhibits the highest value, irrespective of the model architecture. Furthermore, TGCN-(5,5) model is more robust to the consistent and valid perturbations generated by the adaptive attacks as compared to the other two models. Additionally, we note that the predicted inflow state is more affected by the \emph{CVPR} perturbations when the originally predicted inflow state is relatively small.

Interestingly, we note that the predicted inflow states for the adversarially perturbed inputs are, in general, highly correlated---either positively (Fig.~\ref{fig:visualizing_outputs}(b,c) or negatively (FGSM-(0.1,1) and CVPR-(0,1,500) on MLP-5 in Fig.~\ref{fig:visualizing_outputs}(a)---with the originally predicted inflow states, for all the models considered in this experiment. Based on these observations, we conjecture that the crowd-flow prediction models have limited expressiveness---the models are incapable to produce certain outputs irrespective of the inputs.


\begin{figure*}
    \centering
    \begin{subfigure}{0.32\linewidth}
        \centering
        \includegraphics[width=1\linewidth, page=1]{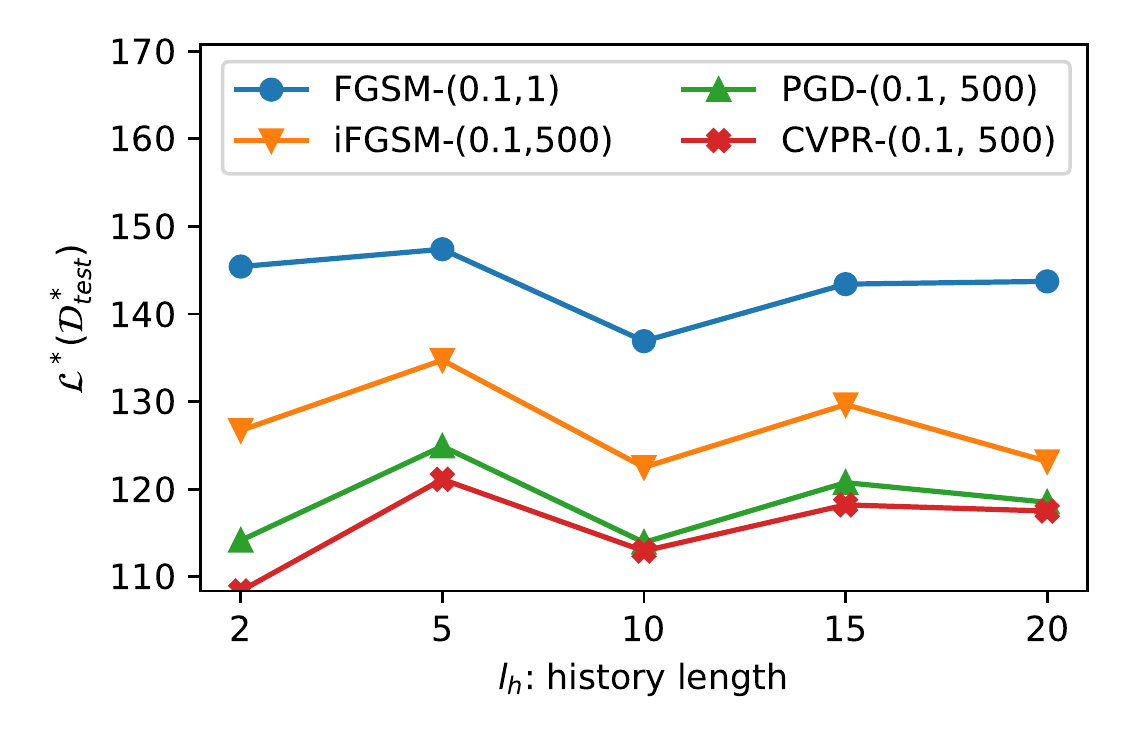}
        \caption{MLP-5 model}
    \end{subfigure}
    \begin{subfigure}{0.32\linewidth}
        \centering
        \includegraphics[width=1\linewidth, page=2]{Figures/adaptive_history.pdf}
        \caption{TGCN-(5,5) model}
    \end{subfigure}
    \begin{subfigure}{0.32\linewidth}
        \centering
        \includegraphics[width=1\linewidth, page=3]{Figures/adaptive_history.pdf}
        \caption{STResnet-2 model}
    \end{subfigure}
    \caption{A comparison of \textbf{the adversarial loss}, $\mathcal{L}^*(\mathcal{D}^*_{test})$, over \textbf{the perturbed test inputs} generated by different attacks for different models (of varying architectures) trained for different history length, $h$. \settings{Dataset is TaxiBJ-16; $\epsilon$ is 0.1}. \take{Typically, when the input history length is increased, the $\mathcal{L}^*(\mathcal{D}^*_{test})$ slightly increases indicating that the models trained on a larger history length are slightly more robust to the adversarial perturbations.}}
    \label{fig:lh_adv_loss}
\end{figure*}

\subsection{Effect of History Length, $h$, on Adversarial Loss}
We analyze the effect of changing the history length, $h$, on the adversarial loss of different attacks. We train three different models---MLP-5, TGCN-(5,5) and STResnet-2---for different history lengths, $h \in \{2, 5, 10, 15, 20\}$ and attack the models using the adversarial attacks considered in this paper. Note that for each $h$, we train a new model as the input of the models trained for different $h$ values are incompatible with each other. For this experiment, we set $\epsilon$=0.1 and $N$=500 for all the attacks. Fig.~\ref{fig:lh_adv_loss}(a-c) reports $\mathcal{L}^*(\mathcal{D}^*_{test})$ values of attacks on the aforementioned three models respectively.

We observe no strict relationship between the adversarial robustness of the models and the history length that they are trained for. However, we note that typically increasing the history length of the model is likely to make the model slightly more robust to the adversarial perturbations, which appears counter-intuitive as a greater history length allows an attacker to add more perturbations to the input. However, recalling what we observed in Section~\ref{sec:results_performance}, we attribute the increasing robustness to the decreasing performance of the models when the history length is increased, which may in turn be due to the accuracy-robustness tradeoff~\cite{ali2021all, wu2021wider}.

\begin{figure*}
    \centering
    \begin{subfigure}{0.32\linewidth}
        \centering
        \includegraphics[width=1\linewidth, page=1]{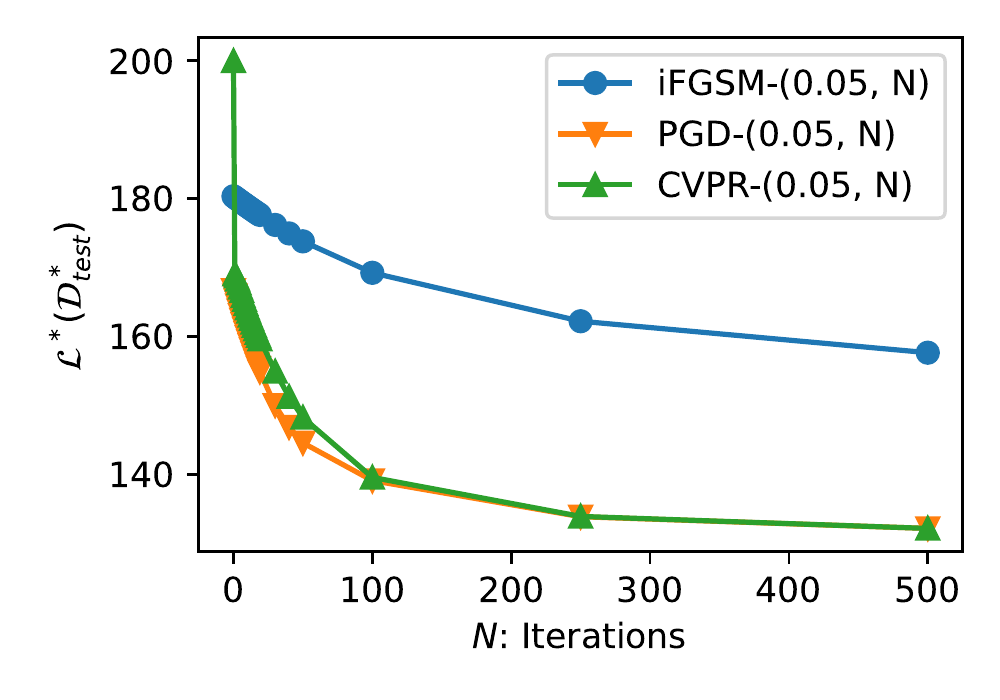}
        \caption{MLP-5 model}
    \end{subfigure}
    \begin{subfigure}{0.32\linewidth}
        \centering
        \includegraphics[width=1\linewidth, page=2]{Figures/attack_results_upa_iterations_mini_adaptive.pdf}
        \caption{TGCN-(5,5) model}
    \end{subfigure}
    \begin{subfigure}{0.32\linewidth}
        \centering
        \includegraphics[width=1\linewidth, page=3]{Figures/attack_results_upa_iterations_mini_adaptive.pdf}
        \caption{STResnet-2 model}
    \end{subfigure}
    \caption{Comparing the decline of adversarial loss, $\mathcal{L}^*(\mathcal{D}^*_{test})$, over the perturbed dataset, $\mathcal{D}^*_{test}$, by different attack algorithms for different model architectures as the attack progresses  assuming \textbf{a D-WB-adaptive attacker}. \settings{Dataset is TaxiBJ-16; $h$ is 5; $\epsilon$ is 0.05}. \take{\emph{CVPR} attack consistently outperforms other attacks in terms of the adversarial loss and speed, given a perturbation budget.}}
    \label{fig:epsilon_adv_iterations}
\end{figure*}

\subsection{Speed of Adversarial Attacks}
Fig.~\ref{fig:epsilon_adv_iterations}(a-c) compares the speed of different attacks to optimize the adversarial loss along the number of iterations for three models---MLP-5, TGCN-(5,5) and STResnet-2---of different architectures, respectively. As the attacks progress, the generated adversarial perturbations become better by each subsequent iteration indicated by the decreasing adversarial loss, irrespective of the attack, which is expected behavior. \emph{CVPR} attack consistently outperforms the other two attacks with a significant margin, specifically notable for TGCN-(5,5) and STResnet-2 models. We attribute this to the explicit induction of crowd-flow validity priors in the perturbation-generating mechanism of \emph{CVPR} attack to generate consistent and valid adversarial perturbations so that the optimizer is not constrained by the consistency and invalidity scores as in eq-\eqref{eq:aware_attack}.




\begin{figure*}
    \centering
    \begin{subfigure}{0.32\linewidth}
        \centering
        \includegraphics[width=1\linewidth, page=1]{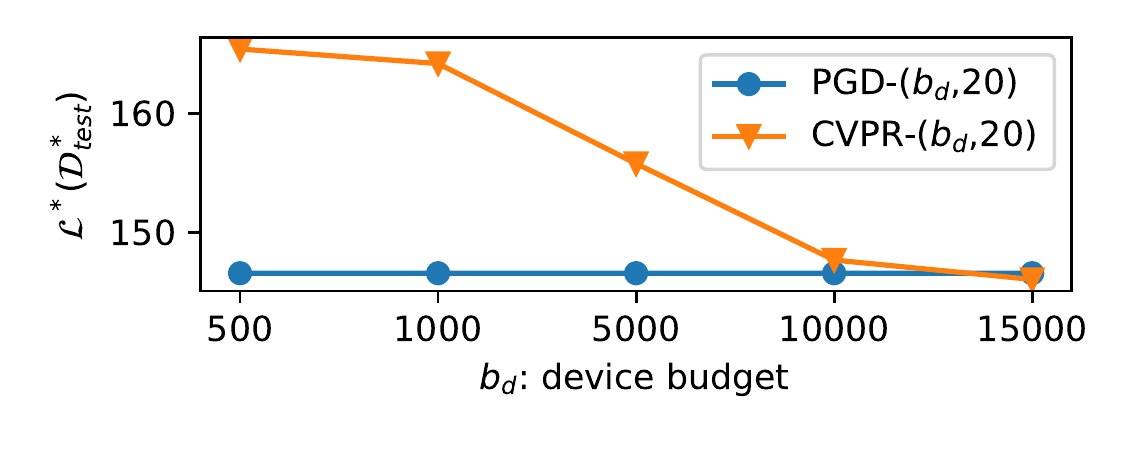}
        \includegraphics[width=1\linewidth, page=1]{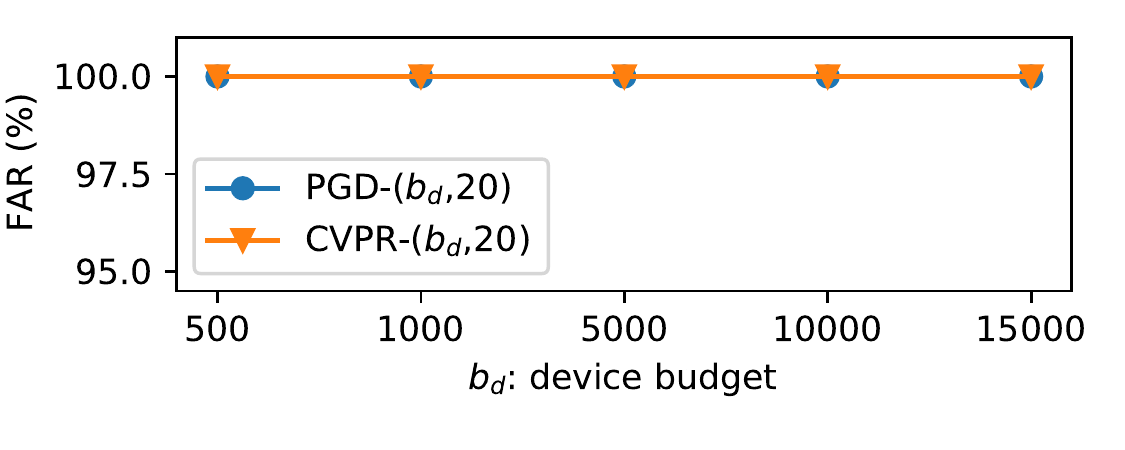}
        \caption{MLP-5 model}
    \end{subfigure}
    \begin{subfigure}{0.32\linewidth}
        \centering
        \includegraphics[width=1\linewidth, page=2]{Figures/_loss.pdf}
        \includegraphics[width=1\linewidth, page=2]{Figures/_validity.pdf}
        \caption{TGCN-(5,5) model}
    \end{subfigure}
    \begin{subfigure}{0.32\linewidth}
        \centering
        \includegraphics[width=1\linewidth, page=3]{Figures/_loss.pdf}
        \includegraphics[width=1\linewidth, page=3]{Figures/_validity.pdf}
        \caption{STResnet-2 model}
    \end{subfigure}
    \caption{Comparing the physical plausibility of the PGD attack and the \emph{CVPR} attack for different model architectures at different device budgets in terms of the adversarial loss, $\mathcal{L}^*(\mathcal{D}^*_{test})$, and FAR of \emph{CaV-detect} assuming \textbf{a D-WB-adaptive attacker}. \settings{Dataset is TaxiBJ-16; $h$ is 5; $b_d$ is the maximum number of devices physically controllable by the attacker}.}
    \label{fig:physical_attacks}
\end{figure*}

\subsection{On Physical-Realizability of Adversarial Attacks}
To further understand the gravity of the adversarial attacks, in this experiment, we study the effectiveness of adversarial attacks under a strictly limited threat model assuming a physical attacker who can read the model weights and inputs but cannot perturb the inputs to the model. P-WB-adaptive attack in Fig.~\ref{fig:threat_models} illustrates such a threat model. In addition, we further limit our attacker by assuming a limited query setting~\cite{khalid2020fadec} that limits our attacker to be only able to query the model 20 times at maximum. We further assume that our attacker has a limited device budget, $b_d$, defining the number of devices, that we refer to as the adversarial devices, which our attacker can physically control. The goal of our attacker is to fool the crowd-prediction model by physically moving the adversarial devices (to simulate adversarial perturbations). We vary the $b_d \in \{500, 1000, 5000, 10000, 15000\}$, and report the adversarial loss of two attacks---PGD-($b_d$, 20) and CVPR-($b_d$, 20)---in Fig.~\ref{fig:physical_attacks} for different values of $b_d$.

Interestingly, we note that for the relatively smaller device budgets, $b_d$, the PGD attack consistently outperforms the \emph{CVPR} attack in terms of $\mathcal{L}^*(\mathcal{D}^*_{test})$ values for all the three models of different architectures considered in this experiment. We attribute this to two reasons. Firstly, the adversarial perturbations generated by PGD attack are relatively more invalid as compared to those generated by the \emph{CVPR} attack, observable in the last row of Fig.~\ref{fig:physical_attacks}, which reports FAR of the \textit{CaV-detect} mechanism. Secondly, the outflow perturbation generating mechanism proposed in eq-\eqref{eq:cvpr_delta_out_generation} implicitly imposes additional constraints on $\boldsymbol{\updelta}_{in}$ and $\boldsymbol{\updelta}_{out}$. While these additional constraints are helpful in the long run, particularly for the stronger adversaries (as observed previously in Fig.~\ref{fig:epsilon_adv_iterations}), they might hurt the attack performance when the perturbation budget is too limited.

Compared to $\mathcal{L}^*(\mathcal{D}^*_{test})$ values in Fig.~\ref{fig:epsilon_adv_loss_adaptive}, $\mathcal{L}^*(\mathcal{D}^*_{test})$ in Fig.~\ref{fig:physical_attacks} are notable smaller, which can simply be attributed to the limited query budget and device budget of the attacker. This shows that although the crowd-flow prediction models are vulnerable to consistent and valid adversarial perturbations under physical settings, realizing the targeted outputs is considerably more challenging than the digital attack settings.

\begin{figure*}
    \centering
    \begin{subfigure}{0.48\linewidth}
        \centering
        \includegraphics[width=0.325\linewidth, page=1]{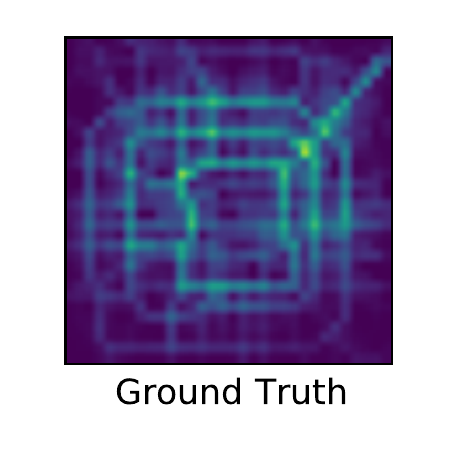}
        \includegraphics[width=0.325\linewidth, page=7]{Figures/st_resnet_expressiveness.pdf}
        \includegraphics[width=0.325\linewidth, page=7]{Figures/st_resnet_expressiveness_2.pdf}
        \caption{Illustrating the ground truth inflow state (recorded in the future), and the adversarial target inflow states.}
    \end{subfigure}
    \begin{subfigure}{0.48\linewidth}
        \centering
        \includegraphics[width=0.325\linewidth, page=2]{Figures/mlp_expressiveness.pdf}
        \includegraphics[width=0.325\linewidth, page=2]{Figures/tgcn_expressiveness.pdf}
        \includegraphics[width=0.325\linewidth, page=2]{Figures/st_resnet_expressiveness.pdf}
        \caption{Comparing the inflow states predicted by three models---MLP-5, TGCN-(5,5), and STResnet-2---over the original inputs.}
    \end{subfigure}
    
    \begin{subfigure}{0.48\linewidth}
        \centering
        \includegraphics[width=0.325\linewidth, page=5]{Figures/mlp_expressiveness.pdf}
        \includegraphics[width=0.325\linewidth, page=5]{Figures/tgcn_expressiveness.pdf}
        \includegraphics[width=0.325\linewidth, page=5]{Figures/st_resnet_expressiveness.pdf}
        \caption{Comparing the inflow states predicted by three models over the PGD-blind attacked inputs optimized for Target-1.}
    \end{subfigure}
    \begin{subfigure}{0.48\linewidth}
        \centering
        \includegraphics[width=0.325\linewidth, page=5]{Figures/mlp_expressiveness_2.pdf}
        \includegraphics[width=0.325\linewidth, page=5]{Figures/tgcn_expressiveness_2.pdf}
        \includegraphics[width=0.325\linewidth, page=5]{Figures/st_resnet_expressiveness_2.pdf}
        \caption{Comparing the inflow states predicted by three models over the PGD-blind attacked inputs optimized for Target-2.}
    \end{subfigure}
    
    \caption{Illustrating limited \emph{expressiveness} (the ability of a model to produce the desired output given an infinite control over the input) of crowd-flow prediction models of different architectures. \settings{Dataset is TaxiBJ-16; $h$ is 5; Attack is PGD-(1,500); $\epsilon$ is 1---indicating infinite control over the inputs}. \take{STResnet-2 model is the most expressive of the three models considered in this experiment, while MLP-5 model is the least expressive}.}
    \label{fig:limited_expressiveness}
\end{figure*}

\subsection{Limited Expressiveness of Crowd-flow Prediction Models}\label{sec:discussion_limitedExpressiveness}
In this experiment, we show that the three crowd-flow prediction models---MLP-5, TGCN-(5,5), and STResnet-2---exhibit limited expressiveness. We define expressiveness as the ability of a model to produce a certain output given an appropriate input. Fig~\ref{fig:limited_expressiveness}(b) shows that all the three models give predictions that are close to the ground truth output (Fig.~\ref{fig:limited_expressiveness}(a)) recorded in the future. To show the limited expressiveness of the models, we assume a strong PGD-(1,500) adversary with $\epsilon=1$ so that the adversary can make any change to the input with \emph{CaV-detect}-blind threat model so that the adversary only has to optimize the inputs and not care about the \emph{CaV-detect} mechanism. Additionally, we assign two adversarial target states---Target-1 and Target-2 in Fig.~\ref{fig:limited_expressiveness}(a)---for the adversary to perturb the inputs (as much as the adversary can) in order to produce the target states at the models' outputs. Fig.~\ref{fig:limited_expressiveness}(c,d) report the output predictions of the aforementioned models when the adversarially perturbed inputs, generated by PGD-(1,500) attack for Target-1 and Target-2 respectively, are fed into the models.

Ideally, one would assume that because an adversary can completely control the input, the adversary can manipulate the model into producing any desired output. However, results reported in Fig.~\ref{fig:limited_expressiveness}(c,d) show that this is not the case with the crowd-flow prediction models. For example, the outputs of the MLP-5 model are significantly different from the targets, which concludes that the MLP-5 model is incapable of producing the target outputs. We say that the MLP-5 model has very limited expressiveness. Although TGCN-(5,5) and STResnet-2 models get significantly closer to the target outputs as compared to the MLP-5 model, they still lack sufficient expressiveness to exactly produce the target output. We attribute this to the mostly clustered and highly similar outputs in the TaxiBJ dataset. Overall, we observe that the STResnet-2 model is the most expressive of the three, which also explains why STResnet models are adversarially less robust compared to TGCN models (as observed in Fig.~\ref{fig:epsilon_adv_loss_adaptive}).

Interestingly, the adversary is only able to get similar output from the MLP-5 model for both the target states. We attribute this to the simplicity of MLP-5 architecture, which fails to capture meaningful and generalizable crowd-flow patterns from the dataset.

\section{Conclusions}
In this paper we studied the adversarial vulnerabilities of the crowd-flow prediction models of three different architectures---Multi-Layer Perceptron, Temporal Graph Convolution Neural Network and Spatio-Temporal ResNet. We extensively analyze the effects of changing the model complexity and crowd-flow data history length on the performance and the adversarial robustness of the resulting models, and find that the crowd-flow prediction models, like other deep learning models, are significantly vulnerable to adversarial attacks. Secondly, we identified and normalized two novel properties---consistency and validity---of the crowd-flow inputs that can be used to detect adversarially perturbed inputs. We therefore propose \emph{CaV-detect} that can detect adversarial inputs with FAR of 0\% by analyzing their consistency and validity---a model input is considered unperturbed if it is both consistent and valid. We then adaptively modify the standard adversarial attacks to evade \emph{CaV-detect} with an FAR of $\sim$80-100\%. Finally, by encoding the consistency and validity priors in the adversarial perturbation generating mechanism, we propose \emph{CVPR} attack, a consistent, valid and physically-realizable adversarial attack that outperforms the adaptive standard attacks in terms of both the target adversarial loss and the FAR of \emph{CaV-detect}. Lastly, insightfully discuss the adversarial attacks on crowd-flow prediction models and show that crowd-flow prediction models exhibit limited expressiveness and can be physically realized by simulating universal perturbations.

\section*{Acknowledgment}
This publication was made possible by NPRP grant \# [13S-0206-200273] from the Qatar National Research Fund (a member of Qatar Foundation). The statements made herein are solely the responsibility of the authors. 

\bibliographystyle{IEEEtran}
\bibliography{Bibliography/ref.bib, Bibliography/references.bib, Bibliography/references2.bib}

\end{document}